\documentclass[10pt,twocolumn,letterpaper]{article}

\usepackage[pagenumbers]{cvpr} 

\usepackage{graphicx}
\usepackage{amsmath}
\usepackage{amssymb}
\usepackage{booktabs}
\usepackage{arydshln}
\usepackage{listings}

\usepackage[pagebackref,breaklinks,colorlinks]{hyperref}

\newlength\savewidth\newcommand\shline{\noalign{\global\savewidth\arrayrulewidth
  \global\arrayrulewidth 1pt}\hline\noalign{\global\arrayrulewidth\savewidth}}
\usepackage[capitalize]{cleveref}
\crefname{section}{Sec.}{Secs.}
\Crefname{section}{Section}{Sections}
\Crefname{table}{Table}{Tables}
\crefname{table}{Tab.}{Tabs.}

\def\cvprPaperID{***} 
\def\confName{CVPR}
\def\confYear{2023}

\begin{document}

\title{HCFormer: Unified Image Segmentation with Hierarchical Clustering}

\author{Teppei Suzuki \vspace{.5em}\\
Denso IT Laboratory, Inc.\\
}
\maketitle

\begin{abstract}
Hierarchical clustering is an effective and efficient approach widely used for classical image segmentation methods.
However, many existing methods using neural networks generate segmentation masks directly from per-pixel features, complicating the architecture design and degrading the interpretability.
In this work, we propose a simpler, more interpretable architecture, called \textit{HCFormer}.
HCFormer accomplishes image segmentation by bottom-up hierarchical clustering and allows us to interpret, visualize, and evaluate the intermediate results as hierarchical clustering results.
HCFormer can address semantic, instance, and panoptic segmentation with the same architecture because the pixel clustering is a common approach for various image segmentation tasks.
In experiments, HCFormer achieves comparable or superior segmentation accuracy compared to baseline methods on semantic segmentation (55.5 mIoU on ADE20K), instance segmentation (47.1 AP on COCO), and panoptic segmentation (55.7 PQ on COCO).\footnote{The code will be publicly available.}
\end{abstract}

\section{Introduction}


Recently proposed image segmentation methods are basically built on neural networks, including convolutional neural networks~\cite{fcn} and transformers~\cite{transformer,vit}, and generate segmentation masks directly from per-pixel features.
However, classical image segmentation methods often use a hierarchical approaches (\eg, hierarchical clustering)~\cite{couprie2013indoor,farabet2012learning,etps,selective-search,ren2003learning}, and the hierarchical strategy improves segmentation accuracy and computational efficiency.
Nonetheless, the neural network-based approaches do not adopt the hierarchical approach, complicating the architecture design and degrading the interpretability.
Therefore, we investigate simpler, more interpretable pipelines for image segmentation from the hierarchical clustering perspective.
\begin{figure}
    \centering
    \includegraphics[clip,width=1\hsize]{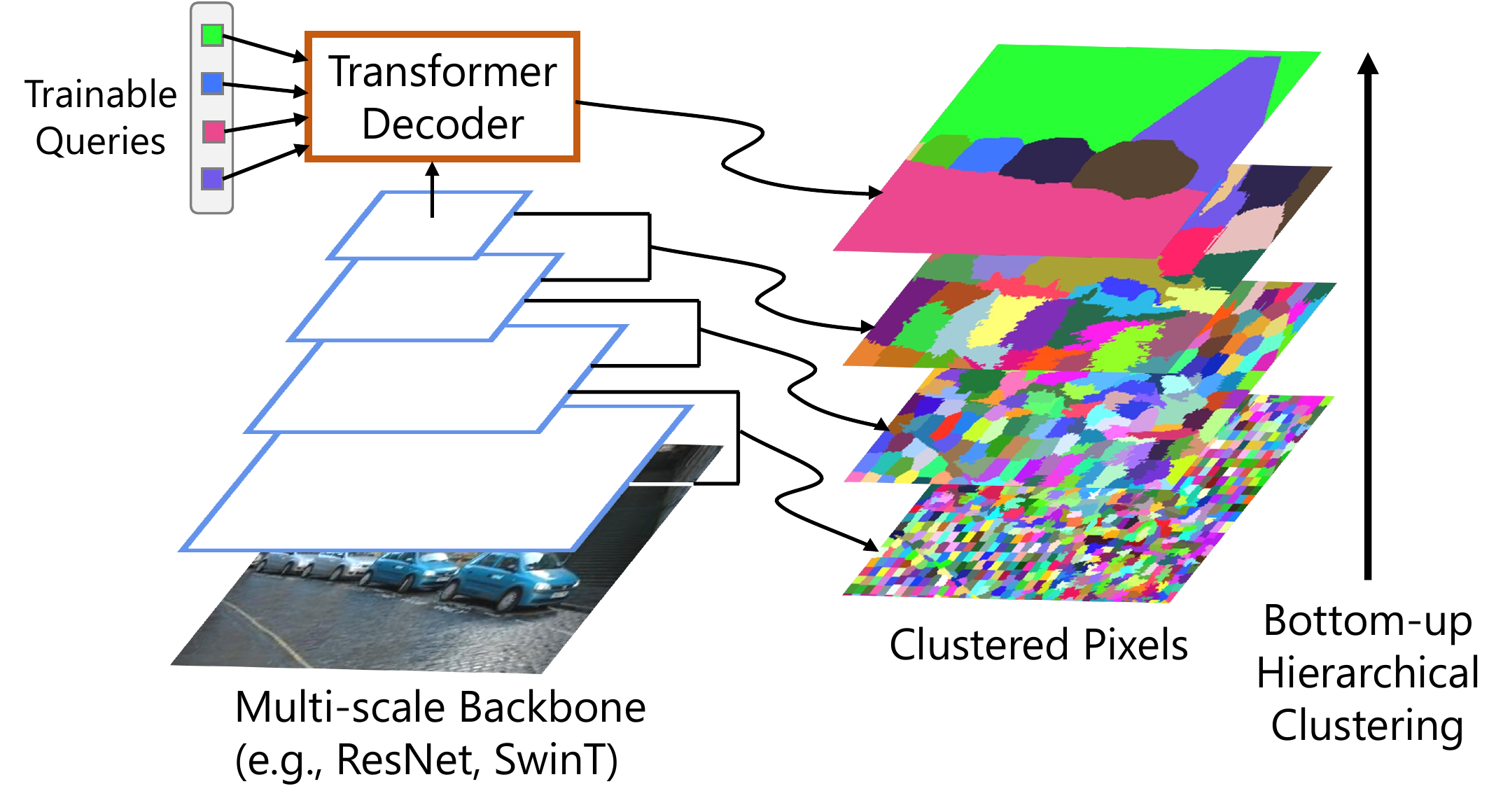}
    \caption{Overview of HCFormer. HCFormer hierarchically groups pixels at downsampling layers in the backbone and then groups clusters obtained from the backbone into an arbitrary number of clusters in the segmentation head.}
    \label{fig:overview}
\end{figure}

In general, segmentation models using neural networks consist of three parts: (i) a backbone model that extracts meaningful features from raw pixels, (ii) a pixel decoder that recovers the spatial resolution lost in the backbone, and (iii) a segmentation head that generates segmentation masks by a classifier or a cross-attention (or cross-correlation) between queries (or kernels) and feature maps.
The pixel decoder is, unlike the others, an inherently unnecessary module to approach image segmentation; in fact, some methods do not adopt it~\cite{fcn,pspnet,deeplabv3}.
Nonetheless, the pixel decoder is used in many segmentation models because the models require per-pixel features to generate high-resolution masks.
The problem is that the decoding is performed in the high-dimensional feature space, and the high dimensionality makes it difficult to interpret, visualize, and evaluate the decoding process, which is one of the causes of decreasing the interpretability of the segmentation models.

We build the segmentation model based on the hierarchical clustering strategy.
Clustering for image segmentation is to group pixels that have the same semantics or ground-truth labels.
We can view it as a special case of downsampling if clustering assigns a representative value (\eg, a class label or a feature vector representing the cluster) to each cluster.
In particular, if elements are grouped based on a fixed window and a representative value is sampled from each group, it is the same as a downsampling scheme.
From this perspective, we accomplish image segmentation without the pixel decoder before the segmentation head.
If we build the backbone model so that its downsampling layers group pixels and sample representative values, the image segmentation can be accomplished as hierarchical clustering, as shown in Fig. \ref{fig:overview}.
This framework simplifies the architecture design and allows us to interpret, visualize, and evaluate intermediate results as clustering results.
Allowing for visualization and evaluation enhances segmentation models' interpretability, that is, the degree to which a human can understand the cause of a decision~\cite{molnar2020interpretable}.

To accomplish the hierarchical clustering in deep neural networks, we propose a clustering module using the attention function~\cite{transformer} as an assignment solver and a model using this module, called \textit{HCFormer}.
HCFormer groups pixels at downsampling layers by the clustering module and realizes image segmentation by hierarchical clustering, as shown in Fig. \ref{fig:overview}.
We attentively design the clustering module to be easily combined with existing backbone models (\eg, ResNet~\cite{resnet} and Swin Transformer~\cite{swin}).
As a result, the hierarchical clustering scheme can be incorporated into the existing backbone models without a change in their feed-forward path.

HCFormer generates segmentation masks by a matrix multiplication between assignment matrices, and the accuracy of this decoding process can be evaluated by some metrics for assessing pixel clustering methods, such as superpixel segmentation~\cite{sp-sota}.
These properties enable an error analysis and provide some architecture-level insight for improving segmentation accuracy, though one may not comprehend why certain decisions or predictions have been made.
For example, when an error occurs at a certain clustering level in HCFormer, at least we know the cause exists in layers before the corresponding downsampling layer in the backbone.
Thus, we may be able to resolve an error by adding layers or modules to the relevant stage in the backbone.
In contrast, it is difficult for the conventional models to evaluate the decoding accuracy because the pixel decoder upsamples pixels in high-dimensional feature space, and the intermediate results are not comparable to the ground-truth labels.
We believe our hierarchical clustering takes the interpretability of segmentation models one step forward, even if it is not a big step.

Since clustering is a common approach for various image segmentation tasks, it can approach many segmentation tasks in the same architecture.
Thus, we evaluate HCFormer on three major segmentation tasks: semantic segmentation (ADE20K~\cite{ade20k} and Cityscapes~\cite{cityscapes}), instance segmentation (COCO~\cite{coco}), and panoptic segmentation (COCO~\cite{coco}).
HCFormer demonstrates comparable or better segmentation accuracy compared to the recently proposed unified segmentation models (\eg, MaskFormer~\cite{mf}, Mask2Former~\cite{m2f}, and K-Net~\cite{knet}) and specialized models for each task, such as Mask R-CNN~\cite{mrcnn}, SOLOv2~\cite{solov2}, SegFormer~\cite{segformer}, CMT-Deeplab~\cite{yu2022cmt}, and Panoptic FCN~\cite{panfcn}.

\section{Method}
We realize hierarchical clustering in deep neural networks by providing a clustering property for downsampling layers.
We may be able to do so by clustering pixels and then sampling representative values from obtained clusters instead of conventional downsampling.
However, the obtained clusters often do not form regular grid structures, and CNN-based backbones do not allow such irregular grid data as input.
Therefore, a straightforward approach, such as downsampling after clustering, is not applicable.

To incorporate the clustering process into existing backbone models while preserving data structures, we propose a \textit{clustering-after-downsampling} strategy.
We show our clustering and decoding pipelines in Fig. \ref{fig:clustering}.
We assume the downsampling used in existing backbone models is cluster-prototype sampling.
Accordingly, we view the pixels in the feature map after downsampling as cluster prototypes and group pixels in the feature map before downsampling.
We first show this clustering process can be realized by the attention~\cite{transformer} in Sec. \ref{sec:attn-as-clst}, and then, we formulate the attention-based clustering module in Sec. \ref{sec:main-claim}.
Finally, we describe our decoding procedure in Sec. \ref{sec:decode}.

\subsection{Clustering as Attention}
\label{sec:attn-as-clst}
We view the attention function~\cite{transformer} from the clustering perspective.
Let $q\in\mathbb{R}^{C\times N_q}$ and $k\in\mathbb{R}^{C\times N_k}$ be a query and a key.
$N_q$ and $N_k$ are the number of tokens for the query and key, and $C$ denotes a feature dimension.
Then, the attention is defined as follows:
\begin{align}
\label{eq:attn}
    \mathrm{Attention}(q,k;s)=\mathrm{Softmax}_\text{row}(q^\top k/s),
\end{align}
where $\top$ denotes the transpose of a matrix and $s$ denotes a scale parameter that is usually defined as $\sqrt{C}$~\cite{vit,transformer}.
$\mathrm{Softmax}_\text{row}(\cdot)$ denotes the row-wise softmax function.
The attention function is generally defined with a query, a key, and a value, but the value is omitted here for simplicity.

When $s\rightarrow0$, eq. \eqref{eq:attn} is equivalent to the following maximization problem:
\begin{align}
    \label{eq:clst-as-attn}
    \underset{A\in \{0,1\}^{N_q\times N_k}}{\arg\max}&\langle A,q^\top k\rangle,\ s.t.,\ \sum_mA_{nm}=1,
\end{align}
where $\langle\cdot,\cdot\rangle$ denotes the Frobenius inner product.
We provide the detailed derivation in Appendix \ref{sec:clst-deriv}.
This maximization problem is interpreted as the clustering problem for $q$ with $k$ as cluster prototypes.
With an inner product as a similarity function, $A_{nm}=1$ if $n$-th query, $q_n$, has the maximum similarity to $m$-th key, $k_m$, among all key tokens; otherwise, $A_{nm}=0$.
In other words, each row of $A$ indicates the index of the cluster assigned to $q_n$, known as the assignment matrix.
Thus, eq. \eqref{eq:clst-as-attn} is an assignment problem, which is a special case or part of the clustering (\eg, agglomerative hierarchical clustering with a hierarchical level of one and one-step $K$-means clustering), and the attention in eq. \eqref{eq:attn} solves the relaxed problem of eq. \eqref{eq:clst-as-attn}.
This insight indicates that we can accomplish clustering after downsampling using the attention in a differentiable form by corresponding the pixels in the feature maps before and after downsampling to queries and keys, respectively.

\subsection{Image Segmentation by Hierarchical Clustering}
\label{sec:main-claim}
\begin{figure}
    \centering
    \includegraphics[clip,width=1\hsize]{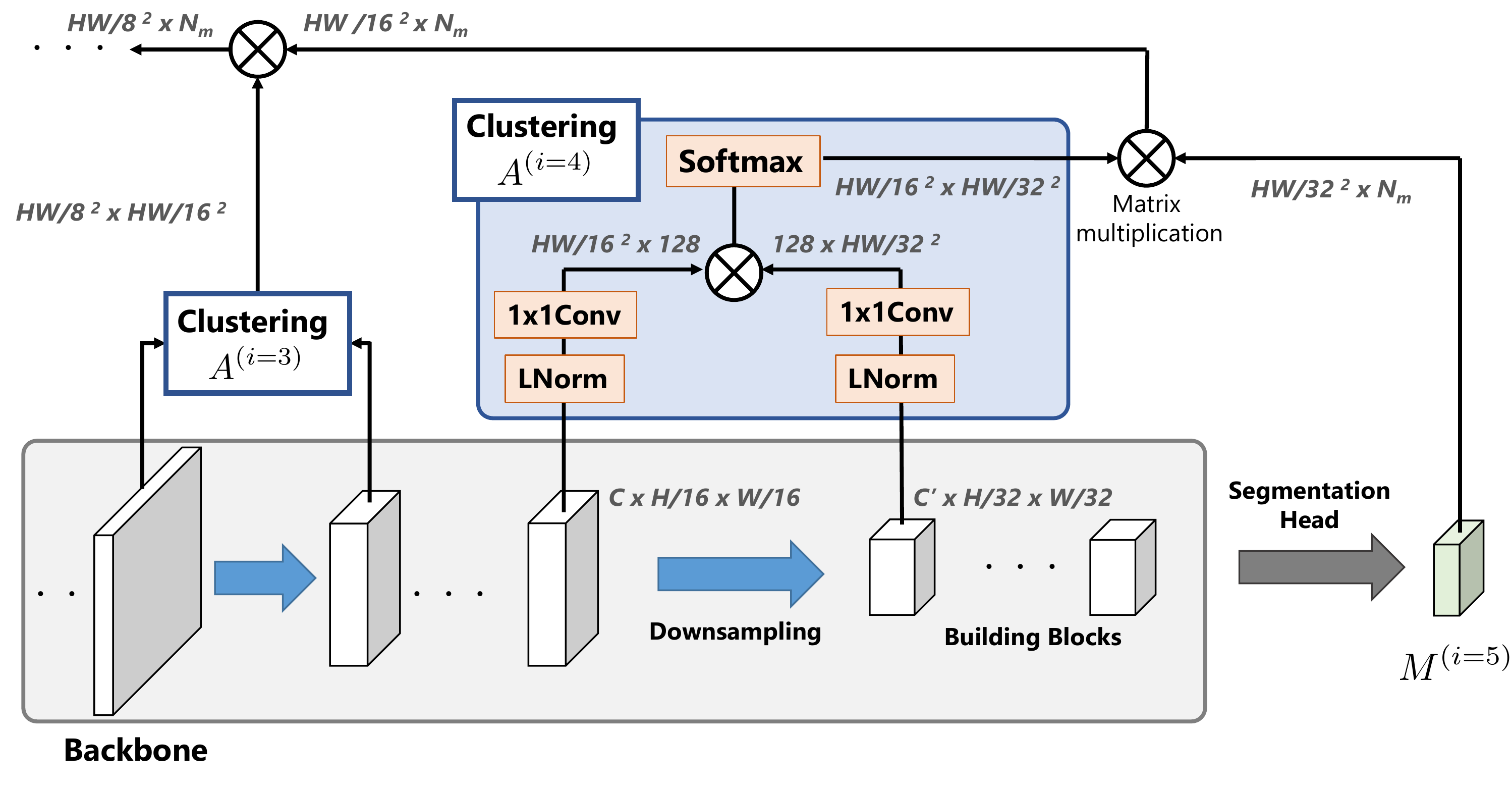}
    \caption{The computational scheme of our clustering module. We can view our clustering module as a variant of the attention module. LNorm and 1$\times$1Conv denote layer normalization~\cite{lnorm} and a convolution layer with a kernel size of 1$\times$1.}
    \label{fig:clustering}
\end{figure}

We show the computational scheme of the proposed clustering module in Fig. \ref{fig:clustering}.
We obtain an intermediate feature map and its downsampled feature map from a backbone model for clustering.
These are fed into layer normalization~\cite{lnorm} and a convolution layer with a kernel size of 1$\times$1, as in the transformer block~\cite{vit,transformer}.
We define the obtained feature maps as $F^{(i)}\in\mathbb{R}^{C_F\times N^{(i)}}$ and $F_d^{(i)}\in\mathbb{R}^{C_F\times N_d^{(i)}}$, where $i\in\mathbb{N}$ denotes a downsampling factor of the spatial resolution that corresponds to the number of applied downsampling layers, and $N^{(i)}$ and $N_d^{(i)}$ denote the number of pixels in the feature maps before and after downsampling (\ie, $N^{(i)}=\frac{H}{2^i}\frac{W}{2^i}$ and $N^{(i)}_d=\frac{H}{2^{i+1}}\frac{W}{2^{i+1}}$, where $H$ and $W$ are the height and width of an input image).
$C_F$ is the number of channels set to 128 in our experiment.
Note that we assume that the downsampling halves height and width respectively, and the $\ell_2$-norm of the feature vector of each pixel is normalized as 1 to make the inner product the cosine similarity.

Then, the proposed clustering is defined as follows:
\begin{align}
    \nonumber
    A^{(i)}&=\mathrm{Clustering}(F^{(i)},F_d^{(i)};s^{(i)})\\ 
    \label{eq:assign}
    &=\mathrm{Softmax}_\text{row}((F^{(i)})^\top F_d^{(i)}/|s^{(i)}|),
\end{align}
where $A^{(i)}\in(0,1)^{N^{(i)}\times N^{(i)}_d}$ is an assignment matrix.
We define a scale parameter, $s^{(i)}\in\mathbb{R}$, as a trainable parameter.
As already described in Sec. \ref{sec:attn-as-clst}, when $s^{(i)}\rightarrow0$, eq. \eqref{eq:assign} corresponds to the clustering problem for $F^{(i)}$ with $F_d^{(i)}$ as cluster prototypes.
Thus, by computing eq. \eqref{eq:assign} at every downsampling layer, HCformer hierarchically groups pixels, as shown in Fig. \ref{fig:overview}.

The obtained feature map from a backbone model is fed into the segmentation head used in \cite{mf} (Fig. \ref{fig:comp}).
Specifically, the feature map is fed into the transformer decoder with trainable queries $Q\in\mathbb{R}^{C_q\times N_m}$, and the mask queries $\mathcal{E}_\text{mask}\in\mathbb{R}^{C_m\times N_m}$ are generated.
The number of queries, $N_m$, is a hyperparameter.
Note that the transformer ``decoder'' differs from the pixel decoder because one of its roles is to generate the mask queries, not to upsample the feature map.
The feature map is also mapped into the $C_m$-dimensional space by a linear layer, and we define them as $\mathcal{E}_\text{feature}^{(i=5)}\in\mathbb{R}^{C_m\times N^{(i)}}$.
Then, the output of the segmentation head is computed as follows:
\begin{align}
    \label{eq:mask}
    M^{(i=5)}=\text{Sigmoid}((\mathcal{E}_\text{feature}^{(i=5)})^\top\cdot\mathcal{E}_\text{mask}).
\end{align}
Note that we assume that the scale $i$ is 5 because the conventional backbone has five downsampling layers.
This segmentation head can be viewed as the clustering, which groups $N^{(i)}$ pixels in the feature map into $N_m$ clusters.\footnote{Eq. \eqref{eq:mask} uses the sigmoid function, not the softmax function.
Thus, eq. \eqref{eq:mask} does not correspond to eq. \eqref{eq:clst-as-attn}.
However, in the post-processing, the mask with the maximum confidence is selected as the prediction, meaning that the pipeline of the segmentation head, including the post-processing, corresponds to eq. \eqref{eq:clst-as-attn}. Further details can be found in Appendix \ref{sec:post-process}.}
Therefore, we also refer to $M^{(i=5)}$ as the assignment matrix.
Unlike the previous work~\cite{mf}, we feed a low-resolution feature map into the segmentation head (eq. \eqref{eq:mask}), which contributes reduction of FLOPs in the segmentation head.

\subsection{Decoding}
\label{sec:decode}
As an output of HCFormer, we obtain the assignment matrices $\{A^{(i)}\}_i$ obtained from the backbone and the output of the segmentation head $M^{(5)}$.
We need to decode them as segmentation masks of an input image for evaluation.

One step of the decoding process is defined as a matrix multiplication between $A^{(i)}$ and $M^{(i+1)}$:
\begin{align}
    \label{eq:decode}
    \nonumber
    M^{(i)}&=A^{(i)}M^{(i+1)}\\
    &=\mathrm{Softmax}_\text{row}\left((F^{(i)})^\top F_d^{(i)}/|s^{(i)}|\right)M^{(i+1)}.
\end{align}
By writing down the calculations in $A^{(i)}$ explicitly, we can notice that it is equivalent to the attention function~\cite{transformer}, although queries, keys, and values are obtained from different layers.
The segmentation mask (\ie, the correspondence between pixels in an input image and the mask queries) is computed by multiplying all $\{A^{(i)}\}_i$ by $M^{(5)}$ as $\prod_{i}A^{(i)}M^{(5)}$.

From the hard clustering perspective, eq. \eqref{eq:clst-as-attn}, we can view multiplying $A^{(i)}$ as copying the cluster's representative value to its elements.
Thus, this decoding process will be accurate if each cluster is composed of pixels that are annotated with the same ground-truth label.
In other words, we can evaluate the accuracy of the decoding from the hard clustering perspective by the undersegmentation error~\cite{sp-sota} that is used for assessing superpixel segmentation.

\subsection{Efficient Computation}
\label{sec:local-attn}
Let $N$ be the number of pixels, and then the computational costs of the proposed clustering are $O(N^2)$, which is the same complexity as the common attention function~\cite{transformer} and is intractable for high-resolution images.
Various studies exist on reducing complexity~\cite{swin,shen2021efficient,bello2021lambdanetworks,xiong2021nystromformer,wang2021pyramid,vaswani2021scaling,qin2022cosformer}, and we approach it with the local attention strategy.

\begin{figure}[t]
    \centering
    \includegraphics[clip,width=1\hsize]{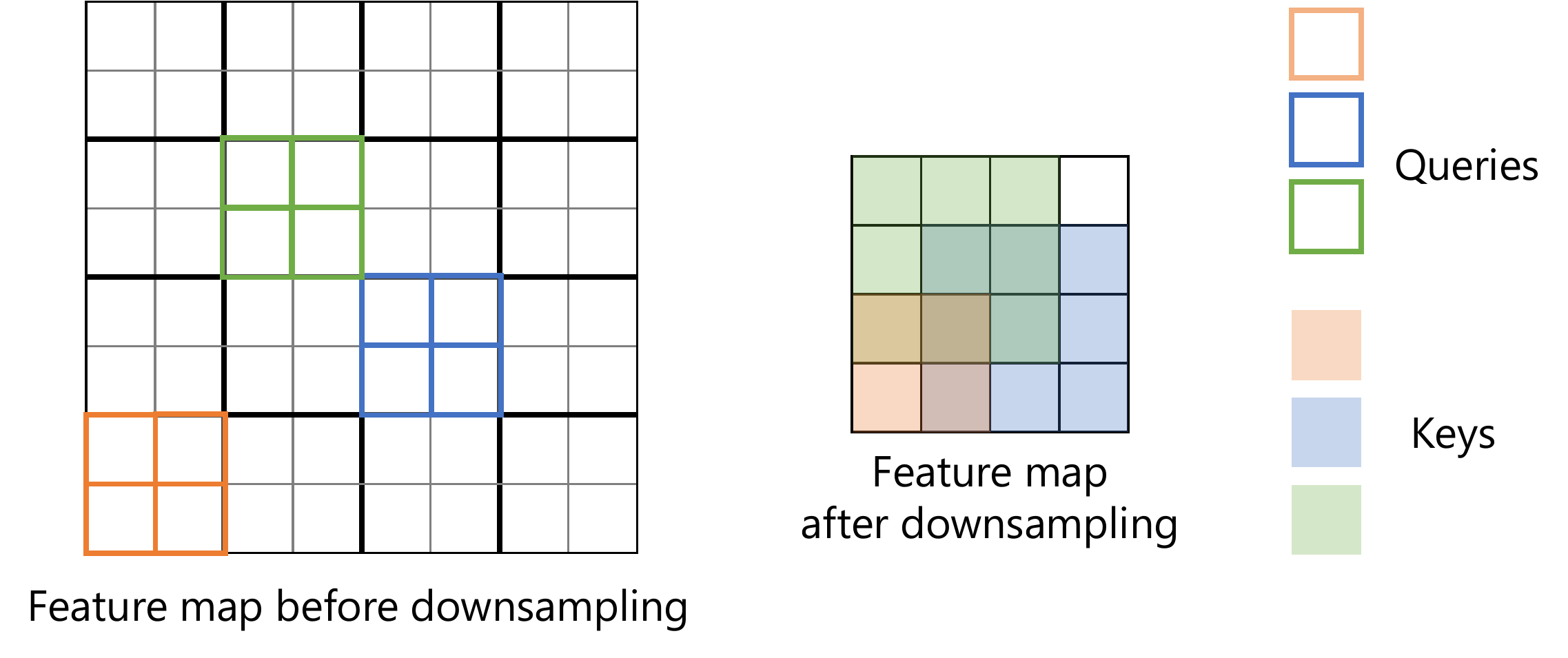}
    \caption{Illustration of the local attention. Each color indicates the correspondence between queries and keys.}
    \label{fig:local-attn}
\end{figure}

We divide the feature map before downsampling, $F^{(i)}$, into 2$\times$2 windows.
The window size is determined by a stride of downsampling, which is assumed to be 2 in this work.
Each window corresponds to the pixel in the feature map after downsampling, $F^{(i)}_d$.
Then, the attention is computed between a pixel in the window and the corresponding pixel in $F^{(i)}_d$ and its surrounding eight pixels, which is the same technique used in superpixel segmentation~\cite{ssn,slic}.
We illustrate this local attention in Fig. \ref{fig:local-attn}.
As a result, the complexity decreases to $O(N)$.
The clustering and decoding processes with this local attention can be easily implemented using deep learning frameworks.
We show an example PyTorch~\cite{pytorch} implementation in Appendix \ref{sec:imple}.

\subsection{Architecture}
\begin{figure}
    \centering
    \begin{tabular}{c}
        \includegraphics[clip,width=0.8\hsize]{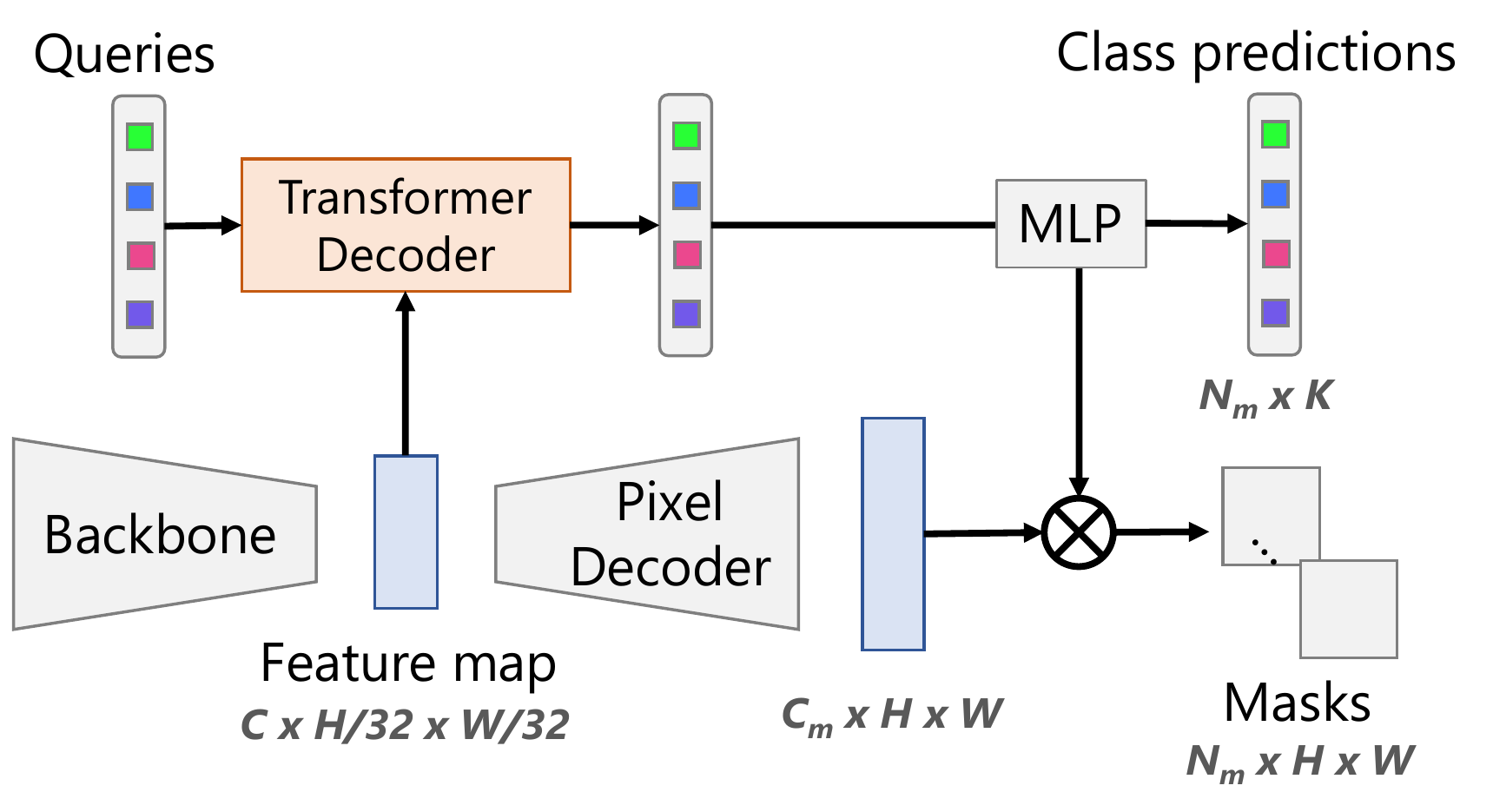} \\
        (a) MaskFormer \\
        \includegraphics[clip,width=0.8\hsize]{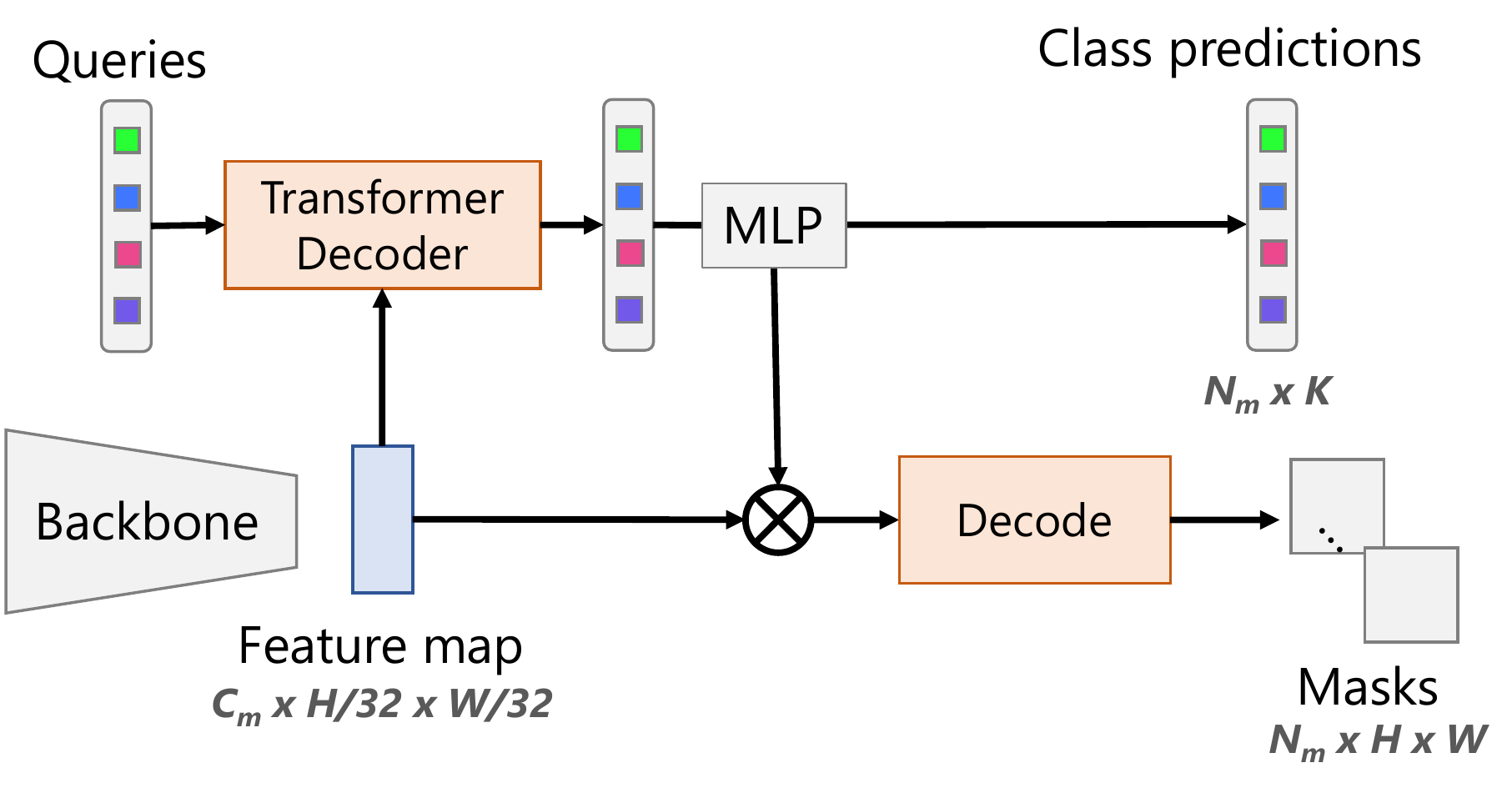}\\
        (b) HCFormer
    \end{tabular}
    \caption{Comparison between MaskFormer~\cite{mf} and HCFormer. MaskFormer upsamples a feature map and then predicts masks. In contrast, HCFormer hierarchically groups pixels in every downsampling layer and the segmentation head and then generates segmentation masks by the decoding defined in eq. \eqref{eq:decode}.}
    \label{fig:comp}
\end{figure}

We show a baseline architecture, MaskFormer~\cite{mf}, and our model using the proposed clustering module in Fig. \ref{fig:comp}.
The proposed model removes the pixel decoder from the baseline architecture, and instead, the decoding block, eq. \eqref{eq:decode}, is integrated for obtaining segmentation masks.
Optionally, we build a six-layer transformer encoder after the backbone, as in MaskFormer~\cite{mf}.
The mask queries are fed into multi-layer perceptrons and classified into $K$ classes.

The flexibility of the downsampling is important for our clustering module because the downsampled pixels are the cluster prototypes that should have representative values of clusters.
However, unlike transformer-based backbones, the receptive field of CNN-based backbones is limited even if the deeper model is used~\cite{erf,segformer}.
Thus, the CNN-based backbones may not sample effective pixels for the clustering module.
To alleviate this problem, we replace downsampling layers to which the proposed clustering module is attached with DCNv2~\cite{dcnv2} with a stride of two, which deforms kernel shapes and modulates weight values.
This replacement is only adopted for a CNN-based backbone.
We verify its effect in Appendix \ref{sec:app-ds}.

\section{Related Work}
\subsection{Image Segmentation with Deep Neural Networks}
Since FCNs~\cite{fcn} have been proposed, deep neural networks have become a de facto standard approach for image segmentation.
As main segmentation tasks, semantic, instance, and panoptic segmentation are studied.

Semantic segmentation is often defined as per-pixel classification tasks, and various FCN-based methods have been proposed~\cite{pspnet,deeplab,segnet,deeplabv3,deeplabv3plus,segformer}.
Instance segmentation has the object detection perspective.
Thus, many methods~\cite{mrcnn,dai2016instance,liang2016reversible,dai2015convolutional,li2017fully} use an object proposal module as the task-specific module, which is used for instance discrimination.
Panoptic segmentation has been proposed in ~\cite{panoptic}, which is a task combining semantic and instance segmentation.
In early work, panoptic segmentation is approached with two separated modules for generating semantic and instance masks and then fusing them~\cite{panoptic,panfpn,pandeeplab,xiong2019upsnet}.
To simplify the framework, Panoptic FCN~\cite{panfcn} unifies the modules by using dynamic kernels.
Panoptic FCN generates kernels from a feature map and produces segmentation masks by cross-correlation between the kernels and feature maps.
As a similar approach, cross-attention with the transformer decoder is adopted in other panoptic segmentation methods~\cite{mdl,detr,mf,m2f,li2022panoptic,yu2022cmt}.
Such approaches also simplify the panoptic segmentation pipeline.

While many studies investigate task-specific modules and architectures, they cannot be applied to other tasks.
Thus, recent work investigates the unified architectures~\cite{mf,knet,m2f}.
These methods use cross-attention-based approaches that can generate segmentation masks regardless of the task definition.
They have demonstrated effectiveness in the various segmentation tasks and achieved comparable or better results than the specialized models.
Our study also focuses on such unified architectures and builds our model based on MaskFormer~\cite{mf}.
Note that, as seen in eq. \eqref{eq:mask}, the segmentation head used in the unified models can be viewed as a clustering module; namely, the unified models would also be clustering-based frameworks.
However, they predict segmentation masks directly from per-pixel features (\ie, not a hierarchical clustering method), and these methods do not allow us to visualize and evaluate the intermediate results.

Several studies define the segmentation tasks as a clustering problem~\cite{kirillov2017instancecut,neven2019instance,kerola2021hierarchical,de2017semantic,kong2018recurrent,yu2022cmt}, especially in the proposal-free instance segmentation methods.
The main focus of these methods is to learn an effective representation for clustering, and their approach is not hierarchical.

Many studies on image segmentation focus on the segmentation head or the pixel decoder.
In particular, various decoder architectures have been proposed, thereby improving segmentation accuracy~\cite{fpn,sharpmask,segnet,unet,fapn,sflow,wojna2017devil,deeplabv3plus}.
The decoder would make the segmentation problem complex: models using the decoder have to solve both the upsampling and pixel classification problems internally.
Such models suffer from an error in the upsampling process of the pixel decoder.
But, it is difficult to know whether the prediction error is due to upsampling or classification because upsampling is performed in the high-dimensional feature space.

Our decoding process can be viewed as an attention-based decoder, as shown in eq. \eqref{eq:decode}, and there are several methods using the attention or similar modules for the pixel decoder~\cite{sflow,pan,lin2019agss,groupvit,fapn}.
However, existing methods do not have a clustering perspective and do not allow us to interpret, visualize, and evaluate the attention map as the clustering results.
In contrast, since HCFormer is built on the clustering perspective, we can interpret and visualize intermediate results as the clustering results and evaluate their accuracy by comparing ground-truth labels.

There are some methods that do not use the pixel decoder.
For example, FCN-32s~\cite{fcn}, the simplest variant of FCNs, does not use the trainable decoder, although the generated masks are low-resolution and the segmentation accuracy is somewhat low.
Several methods~\cite{deeplab,yu2015multi,pspnet,deeplabv3} use dilated convolution~\cite{yu2015multi,deeplab} to keep the resolution in the backbone, which expands the convolution kernel by inserting holes between its consecutive elements and replacing the stride with the dilation rate.
The methods using dilated convolution improve the segmentation accuracy at the expense of FLOPs.

\subsection{Hierarchical Clustering}
Hierarchical clustering approaches are sometimes used to solve image segmentation.
For example, some previous methods~\cite{couprie2013indoor,farabet2012learning,he2015supercnn,tracking-svoxel,spixfcn,etps,spix-conv,suzuki2021implicit,selective-search,ren2003learning} group pixels into small segments using superpixel segmentation~\cite{slic,spix-rim,ssn} as pre-processing; the obtained segments are then merged or classified to obtain the desired cluster.
Such a hierarchical approach often reduces computational costs and improves segmentation accuracy compared to directly clustering or classifying pixels.

A recently proposed method, called GroupViT~\cite{groupvit}, is a bottom-up hierarchical clustering method.
However, GroupViT is specialized in vision transformers~\cite{vit} and semantic segmentation with text supervision.
In contrast, our method can be incorporated into almost all multi-scale backbone models, such as ResNet~\cite{resnet} and Swin Transformer~\cite{swin}, and used for arbitrary image segmentation tasks.

\begin{table*}[t]
    \centering
    \begin{tabular}{l|c|ccc|cc|cc}
        method & backbone & PQ & PQ$^\mathrm{Th}$& PQ$^\mathrm{St}$ & AP$^\mathrm{Th}_\mathrm{pan}$ & mIoU$_\mathrm{pan}$ & \#params. & FLOPS \\ \shline
        Panoptic FCN~\cite{panfcn} & R50 & 44.3 & 50.0 & 35.6 & - & - & - & -\\
        MaskFormer~\cite{mf} & R50    & 46.5 & 51.0 & 39.8 & 33.0 & 57.8 & 45M & 181G \\
        K-Net~\cite{knet} & R50 & 47.1 & 51.7 & 40.3 & - & - & 37M & - \\
        Mask2Former~\cite{m2f} & R50 & 51.9 & 57.7 & 43.0 & 41.7 & 61.7 & 44M & 226G \\ \hdashline
        HCFormer       & R50     & 47.7 & 51.9 & 41.2 & 35.7 & 59.8 & 38M & 87G \\
        HCFormer+     & R50     & 50.2 & 55.1 & 42.8 & 38.4 & 60.2 & 46M & 96G \\ \hline
        MaskFormer~\cite{mf} & Swin-S & 49.7 & 54.4 & 42.6 & 36.1 & 61.3 & 63M & 259G \\
        Mask2Former~\cite{mf} & Swin-S & 54.6 & 60.6 & 45.7 & 44.7 & 64.2 & 69M & 313G \\ \hdashline
        HCFormer       & Swin-S & 50.9 & 55.7 & 43.6 & 38.9 & 63.1 & 62M & 170G \\
        HCFormer+       & Swin-S & 53.0 & 58.1 & 45.3 & 41.2 & 64.4 & 70M & 183G \\ \hline
        CMT-Deeplab~\cite{yu2022cmt} & Axial-R104-RFN & 55.3 & 61.0 & 46.6 & - & - & 270M & 1114G \\
        MaskFormer~\cite{mf} & Swin-L & 52.7 & 58.5 & 44.0 & 40.1 & 64.8 & 212M & 792G \\
        K-Net~\cite{knet} & Swin-L & 54.6 & 60.2 & 46.0 & - & - & - & - \\
        Mask2Former~\cite{m2f} & Swin-L & 57.8 & 64.2 & 48.1 & 48.6 & 67.4 & 216M & 868G\\ \hdashline
        HCFormer       & Swin-L & 55.1 & 60.7 & 46.7 & 44.3 & 66.2 & 210M & 715G \\
        HCFormer+       & Swin-L & 55.7 & 62.0 & 46.1 & 45.3 & 66.4 & 217M & 725G \\
    \end{tabular}
    \caption{Evaluation results for panoptic segmentation with COCO \texttt{val}. HCFormer+ stacks a six-layer transformer encoder after the backbone, as in MaskFormer~\cite{mf}.}
    \label{tab:pq-coco}
\end{table*}
\begin{table*}[t]
    \centering
    \begin{tabular}{c|c|cccc|c|cc}
        method & backbone & AP & AP$^\text{S}$ & AP$^\text{M}$ & AP$^\text{L}$ & AP$^\text{boundary}$ & \#params.  & FLOPs\\ \shline
        Mask R-CNN~\cite{mrcnn} & R50 & 37.2 & 18.6 & 39.5 & 53.3 & 23.1 & 44M & 201G \\
        SOLOv2~\cite{solov2} & R50 & 38.8 & 16.5 & 41.7 & 56.2 & - & - & - \\
        MaskFormer~\cite{mf} & R50 & 34.0 & 16.4 & 37.8 & 54.2 & 23.0 & 45M & 181G \\
        Mask2Former~\cite{m2f} & R50 & 43.7 & 23.4 & 47.2 & 64.8 & 30.6 & 44M & 226G \\ \hdashline
        HCFormer & R50 & 37.4 & 16.7 & 40.0 & 59.7 & 24.6 & 38M & 87G \\
        HCFormer+ & R50 & 40.1 & 18.8 & 43.0 & 62.3 & 26.9 & 46M & 96G \\ \hline
        Mask2Former~\cite{m2f} & Swin-L & 50.1 & 29.9 & 53.9 & 72.1 & 36.2 & 216M & 868G \\ \hdashline
        HCFormer & Swin-L & 46.4 & 24.1 & 50.5 & 70.9 & 32.6 & 210M & 715G \\
        HCFormer+ & Swin-L & 47.1 & 25.6 & 51.5 & 70.3 & 33.3 & 217M & 725G \\
    \end{tabular}
    \caption{Evaluation results for instance segmentation with COCO \texttt{val}. HCFormer+ stacks a six-layer transformer encoder after the backbone.}
    \label{tab:coco-is}
\end{table*}

\section{Experiments}
\subsection{Implementation Details}
We implement our model on the Mask2Former author's implementation.\footnote{\url{https://github.com/facebookresearch/Mask2Former}}
We use the transformer decoder with the masked attention~\cite{m2f}, and the number of decoder layers is set to 8.
Note that we do not use multi-scale feature maps in the transformer decoder because HCFormer does not have the pixel decoder.
Thus, HCFormer is built on top of MaskFormer rather than on top of Mask2Former since the transformer decoder used in HCFormer and MaskFormer is independent of the pixel decoder.
We describe the detailed difference between transformer decoders of HCFormer and Mask2Former in Appendix \ref{sec:diff-m2f}.

We train HCFormer with almost the same protocol as in Mask2Former~\cite{m2f}; we use the binary cross-entropy loss and the dice loss~\cite{dice} as the mask loss: $\mathcal{L}_\text{mask}=\lambda_\text{ce}\mathcal{L}_\text{ce}+\lambda_\text{dice}\mathcal{L}_\text{dice}$.
The training loss combines mask loss, classification loss, and an additional regularization term for $s^{(i)}$ in eq. \eqref{eq:assign}: $\mathcal{L}_\text{mask}+\lambda_\text{cls}\mathcal{L}_\text{cls}+\lambda_\text{reg}\mathcal{L}_\text{reg}$, where $\mathcal{L}_\text{cls}$ is a cross-entropy loss and $\mathcal{L}_\text{reg}=\sum_i|s^{(i)}|$.
$\lambda_\text{reg}$ is set to $0.1$.
The regularization enforces the proposed attention-based clustering, eq. \eqref{eq:assign}, to be the hard clustering, eq. \eqref{eq:clst-as-attn}.
Following \cite{m2f}, we set $\lambda_\text{ce}=5.0$, $\lambda_\text{dice}=5.0$, and $\lambda_\text{cls}=2.0$.
Another training protocol is the same as for Mask2Former~\cite{m2f} (\eg, optimizer, its hyperparameters, and the number of training iterations).
The details can be found in Appendix \ref{sec:detailed-setup}.

We use the same post-processing as in \cite{mf}: we multiply class confidence and mask confidence and use the output as the confidence score.
Then, the mask with the maximum confidence score is selected as the predicted mask.
The details can be found in Appendix \ref{sec:post-process}.

The clustering module defined in eq. \eqref{eq:assign} is incorporated into every downsampling layer, except for those in the stem block of ResNet~\cite{resnet} and the patch embedding layer in Swin Transformer~\cite{swin}.
Thus, the hierarchical level is 3 (\ie, $\{A^{(2)},A^{(3)},A^{(4)}\}$ are computed).
The relation between the hierarchical level and downsampling layers using the clustering module is shown in Fig. \ref{fig:backbone} in the appendix. 

\subsection{Main Results}
\label{sec:main-exp}
As baseline methods, we choose the recently proposed methods for unifying segmentation tasks, MaskFormer~\cite{mf}, K-Net~\cite{knet}, and Mask2Former~\cite{m2f}, and specialized models for each task~\cite{panfcn,yu2022cmt,segformer,mrcnn,solov2}.
We compare our method to the baselines with several backbones, ResNet-50~\cite{resnet}, Swin-S~\cite{swin}, and Swin-L~\cite{swin}.
Note that ResNet-50 and Swin-S were pretrained with ImageNet-1K, and Swin-L was pretrained with ImageNet-22K.
The training procedure and evaluation metrics are following ~\cite{m2f}.
We trained models three times and reported medians.

We first show the evaluation results for panoptic segmentation on the COCO validation dataset~\cite{coco} in Tab. \ref{tab:pq-coco}.
HCFormer outperforms the unified architectures, MaskFormer and K-Net, and Panoptic FCN for all metrics with fewer parameters.
The additional transformer encoder (\ie, HCFormer+) boosts the segmentation accuracy for the relatively small backbone models and outperforms CMT-Deeplab with fewer parameters and FLOPs.
Although the transformer decoder used in HCFormer is different from that used in MaskFormer in this comparison, we verify HCFormer still outperforms MaskFormer even when MaskFormer's transformer decoder is applied (see Appendix \ref{sec:app-ablation}).

Mask2Former shows the best accuracy, though it sacrifices the FLOPs.
According to \cite{m2f}, using multi-scale features in the transformer decoder improves 1.7 PQ for Mask2Former with the ResNet-50 backbone (\ie, PQ of Mask2Former without the multi-scale feature maps is 50.2, which is the same as PQ of HCFormer+ with the ResNet-50 backbone in Tab. \ref{tab:pq-coco}).
Thus, we believe the gap in PQ between HCFormer and Mask2Former is due to the design of the transformer.
In other words, PQ of HCFormer is comparable to that of Mask2Former in the same setting, but HCFormer shows lower FLOPs and has the interpretability.
The detailed analysis is described in Appendix \ref{sec:diff-m2f}.

Tab. \ref{tab:coco-is} shows the average precision on instance segmentation tasks.
HCFormer also outperforms MaskFormer and the specialized model, Mask R-CNN and SOLOv2.
Notably, HCFormer significantly improves AP$^\text{L}$ from MaskFormer, which indicates HCFormer can generate well-aligned masks for large objects.
For recovering the large objects by the pixel decoder, the model needs to capture the long-range dependence, which may be difficult for the simple pixel decoder.
Thus, the conventional pipelines need a well-design decoder to recover the large objects, such as that used in Mask2Former.
However, since HCFormer groups pixels locally and hierarchically, it may not need long-range dependence to capture the large objects, compared to the conventional methods.
As a result, HCFormer outperforms MaskFormer, especially in terms of AP$^\text{L}$.

\begin{figure*}
    \centering
        \includegraphics[clip,width=0.245\hsize]{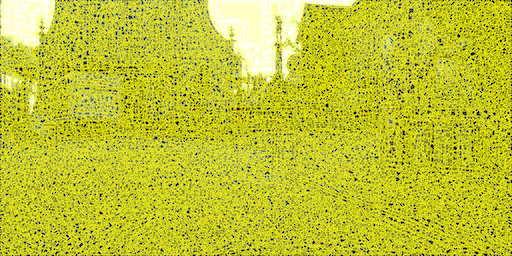}
        \includegraphics[clip,width=0.245\hsize]{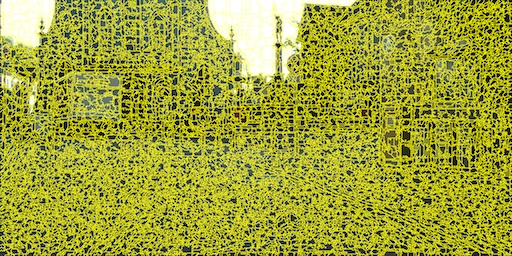} 
        \includegraphics[clip,width=0.245\hsize]{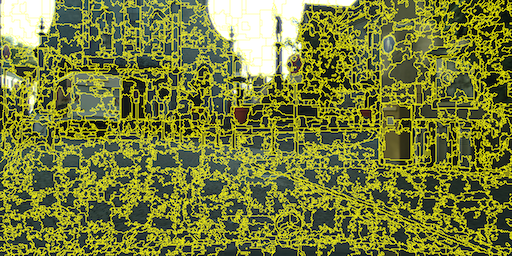}
        \includegraphics[clip,width=0.245\hsize]{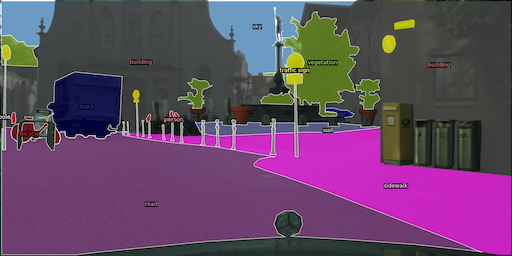}\\
        \includegraphics[clip,width=0.245\hsize]{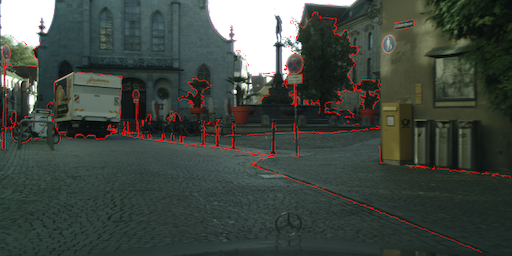} 
        \includegraphics[clip,width=0.245\hsize]{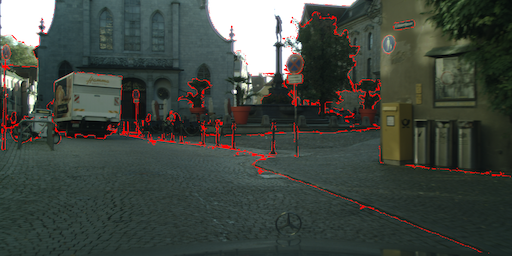} 
        \includegraphics[clip,width=0.245\hsize]{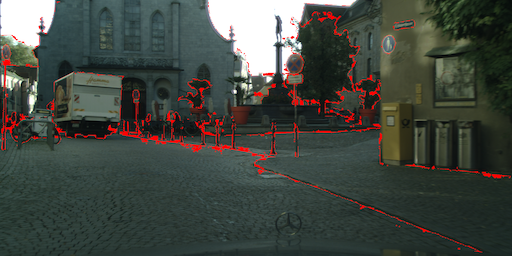} 
        \includegraphics[clip,width=0.245\hsize]{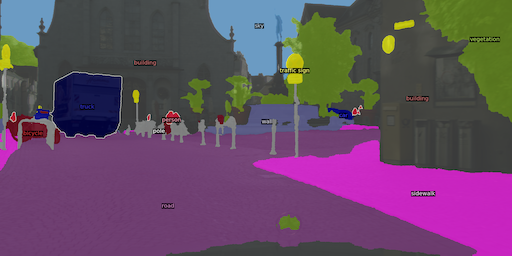}
    \caption{Hierarchical clustering results for Cityscapes. From left to right, the top row shows the cluster boundaries obtained from downsampling layers and ground truth. The bottom row shows undersegmentation errors~\cite{sp-sota} corresponding to the top row results as red regions, which measure the ``leakage'' of clusters for ground truth, and a predicted mask.}
    \label{fig:city-vis}
\end{figure*}

\begin{table}[t]
    \centering
    \scalebox{0.9}{
        \begin{tabular}{c|cc|cc}
            method & backbone & crop size & mIoU & FLOPs\\ \shline
            MaskFormer~\cite{mf} & R50 & 512 & 44.5 & 53G\\
            Mask2Former~\cite{m2f} & R50 & 512 & 47.2 & 73G\\ \hdashline
            HCFormer & R50 & 512 & 45.5 & 29G\\
            HCFormer+ & R50 & 512 & 46.9 & 32G\\ \hline
            SegFormer~\cite{segformer} & MiT-B2 & 512 & 46.5 & 62G \\
            MaskFormer~\cite{mf} & Swin-S & 512 & 49.8 & 79G\\
            Mask2Former~\cite{m2f} & Swin-S & 512 & 51.3 & 98G\\ \hdashline
            HCFormer & Swin-S & 512 & 48.8 & 56G\\
            HCFormer+ & Swin-S & 512 & 50.1 & 58G\\ \hline
            SegFormer~\cite{segformer} & MiT-B5 & 640 & 51.0 & 184G\\
            MaskFormer~\cite{mf} & Swin-L & 640 & 54.1 & 375G\\
            Mask2Former~\cite{m2f} & Swin-L & 640 & 56.1 & 403G \\ \hdashline
            HCFormer & Swin-L & 640 & 55.2 & 338G\\
            HCFormer+ & Swin-L & 640 & 55.5 & 342G
        \end{tabular}
    }
    \caption{Evaluation results for semantic segmentation with ADE20K \texttt{val}. HCFormer+ stacks a six-layer transformer encoder after the backbone.}
    \label{tab:ade20k-ss}
\end{table}
\begin{table}
        \begin{tabular}{c|c|cc}
            method & backbone & mIoU & FLOPs\\ \shline
            MaskFormer~\cite{mf} & R50 & 76.5 & 405G\\
            Mask2Former~\cite{m2f} & R50 & 79.4 & 527G\\ \hdashline
            HCFormer & R50 & 76.7 & 196G\\
            HCFormer+ & R50 & 78.3 & 225G\\ \hline
            SegFormer~\cite{segformer} & MiT-B2 & 81.0 & 717G\\
            MaskFormer~\cite{mf} & Swin-S & 78.5 & 599G\\
            Mask2Former~\cite{m2f} & Swin-S& 82.6 & 727G\\ \hdashline
            HCFormer & Swin-S & 79.3 & 398G\\
            HCFormer+ & Swin-S & 80.0 & 427G\\ \hline
            SegFormer~\cite{segformer} & MiT-B5 & 82.4 & 1460G\\
            MaskFormer~\cite{mf} & Swin-L & 81.8 & 1784G\\
            Mask2Former~\cite{m2f} & Swin-L & 83.3 & 1908G\\ \hdashline
            HCFormer & Swin-L & 81.6 & 1578G\\
            HCFormer+ & Swin-L & 82.0 & 1607G
        \end{tabular}
    \caption{Evaluation results for semantic segmentation with Cityscapes \texttt{val}. HCFormer+ stacks a six-layer transformer encoder after the backbone.}
    \label{tab:city-ss}
\end{table}
Tabs. \ref{tab:ade20k-ss} and \ref{tab:city-ss} show the results on semantic segmentation.
On ADE20k, HCFormer outperforms MaskFormer and the specialized model, SegFormer, and Mask2Former shows the best mIoU.
However, on Cityscapes~\cite{cityscapes} (Tab. \ref{tab:city-ss}), mIoU of HCFormer is worse than that of SegFormer, although HCFormer improves mIoU from MaskFormer.
Unlike COCO and ADE20K, the Cityscapes dataset contains only urban street scenes, and there are many thin or small objects, such as poles, signs, and traffic lights.
MaskFormer and HCFormer do not have specialized modules to capture such small and thin objects; hence such objects would be missed in an intermediate layer.

For further analysis, we visualize intermediate clustering results and a predicted mask using HCFormer with Swin-L in Fig. \ref{fig:city-vis}.
From the visualization of undersegmentation error~\cite{sp-sota}, we find that the model groups pixels well except for the boundaries in intermediate clustering, but it misclassifies clusters for the small objects in this image (\eg, poles and persons in the center).
HCFormer cannot extract semantically discriminative features for such small objects, although the obtained features are discriminative in terms of clustering.
Thus, to solve errors for the small objects, we may improve the transformer decoder or stack more layers on the backbone for the coarsest scale.
We believe that this analysis plays one of the roles in interpretability, and conventional models do not allow for this type of analysis.
We present further results and analysis on other datasets in Appendix \ref{sec:app-vis}.

\subsection{Ablation Study}
To investigate the effect of the hierarchical level and the more effective architecture, we evaluate PQ, FLOPs, and the number of parameters of our model with various hierarchical levels and different numbers of transformer decoder layers.
We use the COCO dataset and the ResNet-50 backbone for evaluation.
The additional ablation study can be found in Appendix \ref{sec:app-ablation}.
\begin{table}[t]
    \centering
    \begin{tabular}{c|cccc}
        Hierarchical Level & 0 & 1 & 2 & 3 \\ \shline
        PQ & 41.9 & 45.6 & 46.7 & 47.7 \\
        FLOPs & 82G & 83G & 84G & 87G \\
        \#Params & 37M & 38M & 38M & 38M 
    \end{tabular}
    \caption{Evaluation results for COCO \texttt{val.} with various hierarchical levels.}
    \label{tab:hlevel}

\end{table}
\begin{table}
    \begin{tabular}{c|ccccc}
        \#Decoder Layers & 1 & 2 & 4 & 8 & 16\\ \shline
        PQ & 42.0 & 44.8 & 46.7 & 47.7 & 48.0\\
        FLOPs & 85G & 85G & 86G & 87G & 90G\\
        \#Params & 27M & 29M & 32M & 38M & 51M
    \end{tabular}
    \caption{Evaluation results for COCO \texttt{val.} with various numbers of transformer decoder layers.}
    \label{tab:num-dlayer}
\end{table}

The results are shown in Tabs. \ref{tab:hlevel} and \ref{tab:num-dlayer}.
The higher the hierarchical level, the higher the resolution of the predicted masks, increasing PQ and FLOPs.
The hierarchical level of 2 would be preferred regarding the balance between accuracy and computational costs in practice, though we adopt the hierarchical level of 3 in Sec. \ref{sec:main-exp}.
In particular, PQ in our model is still comparable to that of MaskFormer~\cite{mf} even when the hierarchical level is 2.

Our model does not work well with only one decoder layer, as with other models using the transformer decoder~\cite{mf,m2f,detr}.
Regarding the balance between accuracy and computational costs, four or eight layers would be suitable for our model.

\section{Limitations}
HCFormer accomplishes the hierarchical clustering framework by removing the pixel decoder before the segmentation head.
However, we do not argue that the pixel decoder is useless, especially in improving segmentation accuracy.
Mask2Former uses the multi-scale feature maps obtained from the pixel decoder in the transformer decoder, significantly improving the segmentation accuracy.
Also, there are many techniques leveraging the pixel decoder to improve segmentation accuracy~\cite{fpn,sflow,segformer,deeplabv3plus}.
Thus, one of the limitations of HCFormer is that it cannot take advantage of existing techniques using the pixel decoder to improve the segmentation accuracy.
This limitation causes the gap between the accuracy of HCFormer and Mask2Former.
We will explore alternatives to the modules leveraging the pixel decoder to improve the accuracy in future work.

\section{Conclusion}
\label{sec:conclusion}
We proposed an attention-based clustering module easily incorporated into existing backbone models, such as ResNet and Swin Transformer.
As a result, we accomplished image segmentation via hierarchical clustering in deep neural networks and simplified the segmentation architecture design by removing the pixel decoder before the segmentation head.
In experiments, we verified that our method achieves comparable or better accuracy than baseline methods for semantic, instance, and panoptic segmentation.
Since our hierarchical clustering allows interpretation, visualization, and evaluation of the intermediate results, we believe that the hierarchical clustering strategy enhances the interpretability of the segmentation models.

{\small
\bibliographystyle{ieee_fullname}
\bibliography{egbib}
}

\newpage

\appendix

\setcounter{table}{0}
\renewcommand{\thetable}{\Roman{table}}
\setcounter{figure}{0}
\renewcommand{\thefigure}{\Roman{figure}}

\section{Detailed derivation of Eq. \eqref{eq:clst-as-attn}}
\label{sec:clst-deriv}
Let $v\in\mathbb{R}^n$ be an $n$-dimensional vector fed into the softmax function, which corresponds to a row of $q^\top k$ in eq. \eqref{eq:attn}, and then $i$-th output of $\mathrm{softmax}(v)$ with scale $s$ is:
\begin{align}
    \label{eq:softmax}
    a_i = \frac{\exp(v_i/s)}{\sum_j\exp(v_j/s)}.
\end{align}
We define $\hat{v}=\max_j v_j$ and modify eq. \eqref{eq:softmax} as follows:
\begin{align}
    \nonumber
    a_i &= \frac{\exp(v_i/s)}{\sum_j\exp(v_j/s)}\cdot\frac{\exp(-\hat{v}/s)}{\exp(-\hat{v}/s)}\\
        &= \frac{\exp((v_i-\hat{v})/s)}{\sum_j\exp((v_j-\hat{v})/s)},
\end{align}
where $v_i-\hat{v} <= 0$.
For numerical calculations, when $s\rightarrow 0$, $\exp((v_i-\hat{v})/s)=\exp(-\infty)=0$ if $v_i\neq\hat{v}$, and $\exp((v_i-\hat{v})/s)=\exp(0/s)=1$ if $v_i=\hat{v}$.
In this case, eq. \eqref{eq:softmax} is equivalent to the following problem:
\begin{align}
    \label{eq:argmax}
    \underset{a\in\{0,1\}^n}{\arg\max}\ a^\top v,\ s.t.,\ \sum_i a_i=1.
\end{align}
The solution of this problem is a one-hot vector that holds $a_l=1,\ l=\arg\max_j v_j$.
By considering eq. \eqref{eq:clst-as-attn} for each row, eq. \eqref{eq:clst-as-attn} is equivalent to eq. \eqref{eq:argmax}.
Hence, when $s\rightarrow0$, eq. \eqref{eq:attn} is equivalent to eq. \eqref{eq:assign}.

\section{Example PyTorch implementation}
\label{sec:imple}
We show example PyTorch~\cite{pytorch} implementation of the proposed clustering and decoding with the local attention in Listings \ref{list:clst} and \ref{list:dec}.
We can easily implement the local attention by using the \texttt{unfold} and \texttt{einsum} functions.
\begin{lstlisting}[language=Python,caption=The proposed soft clustering (eq. \eqref{eq:assign}),label=list:clst]
import torch
import torch.nn.functional as F


def clustering(feat, downsampled_feat, scale):
    batch_size, emb_dim, height, width = feat.shape
    # get 9 candidate clusters and corresponding pixel features
    candidate_clusters = F.unfold(downsampled_feat, kernel_size=3, padding=1).reshape(batch_size, emb_dim, 9, -1)
    feat = F.unfold(feat, kernel_size=2, stride=2).reshape(batch_size, emb_dim, 4, -1)
    # calculate similarities
    similarities = torch.einsum('bkcn,bkpn->bcpn', (candidate_clusters, feat)).reshape(batch_size*9, 4, -1)
    similarities = F.fold(similarities, (height, width), kernel_size=2, stride=2).reshape(batch_size, 9, height, width)
    # normalize
    soft_assignment = (similarities / scale.abs()).softmax(1)
    return soft_assignment
\end{lstlisting}
\begin{lstlisting}[language=Python,caption=The decoding (eq. \eqref{eq:decode}),label=list:dec]
import torch
import torch.nn.functional as F


def decode(x, A):
    batch_size, _, height, width = A.shape
    n_channels = x.shape[1]
    # get 9 candidate clusters and corresponding assignments
    candidate_clusters = F.unfold(x, kernel_size=3, padding=1).reshape(batch_size, n_channels, 9, -1)
    A = F.unfold(A, kernel_size=2, stride=2).reshape(batch_size, 9, 4, -1)
    # decoding
    decoded_features = torch.einsum('bkcn,bcpn->bkpn', (candidate_clusters, A)).reshape(batch_size, n_channels * 4, -1)
    decoded_features = F.fold(decoded_features, (height, width), kernel_size=2, stride=2)
    return decoded_features
\end{lstlisting}

\section{Difference between the transformer decoders of HCFormer and Mask2Former}
\label{sec:diff-m2f}
\begin{figure}[t]
    \centering
    \begin{tabular}{cc}
        \includegraphics[clip,width=0.45\hsize]{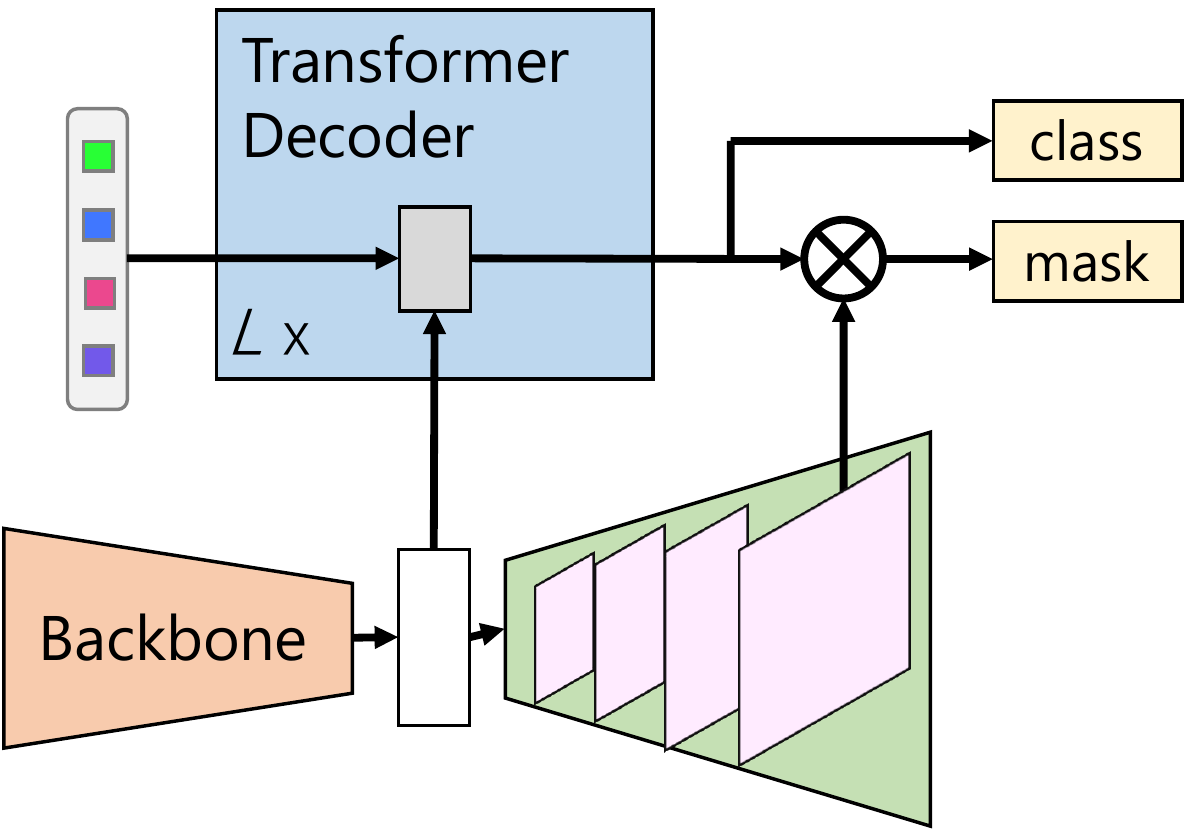} &
        \includegraphics[clip,width=0.45\hsize]{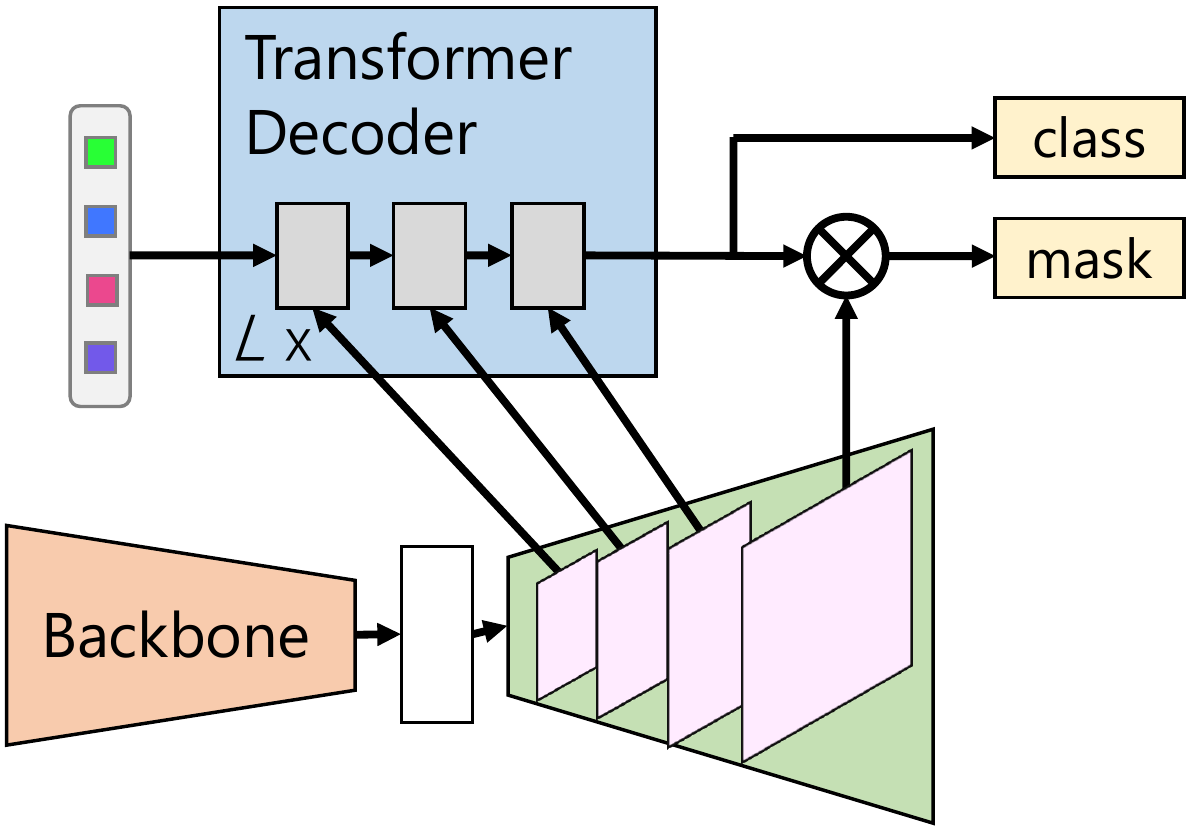} \\
        (a) MaskFormer & (b) Mask2Former 
    \end{tabular}
    \includegraphics[clip,width=0.6\hsize]{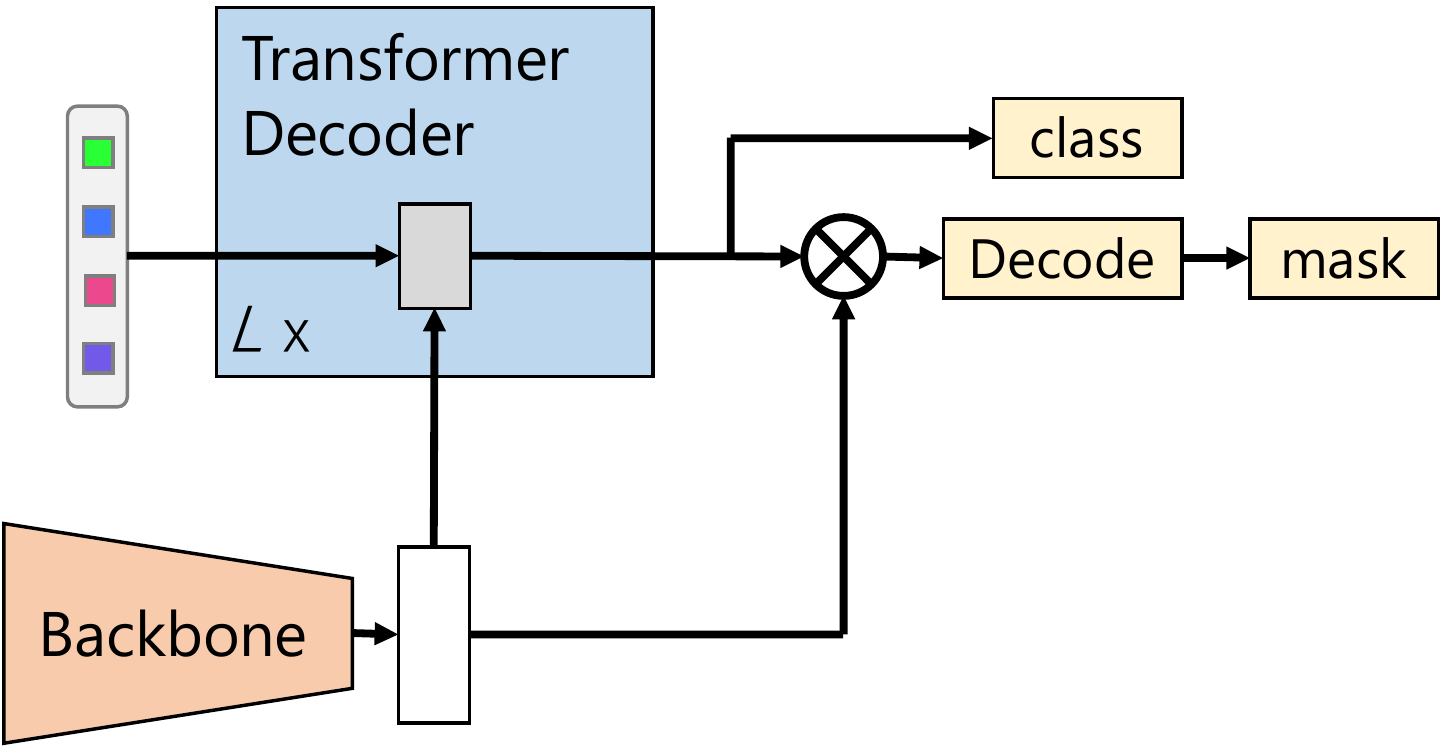} \\
    (c) HCFormer
\caption{The architecture of MaskFormer~\cite{mf}, Mask2Former~\cite{m2f}, and HCFormer.}
    \label{fig:m2f-arc}
\end{figure}

The architectures of MaskFormer~\cite{mf}, Mask2Former~\cite{m2f}, and HCFormer are shown in Fig. \ref{fig:m2f-arc}.
Mask2Former feeds the Multi-scale feature maps obtained from the pixel decoder into the transformer decoder, unlike MaskFormer and HCFormer.
The difference between HCFormer and MaskFormer is only the decoding process, but the difference between HCFormer and Mask2Former is, in addition to the decoding process, the inputs to the transformer decoder.
Therefore, in terms of the architecture design, we believe HCFormer is built on MaskFormer rather than Mask2Former.

In our experiments, the accuracy of HCFormer is inferior to that of Mask2Former.
However, the PQ of HCFormer is comparable to that of Mask2Former without using multi-scale feature maps in the transformer decoder (\ie, the input to the transformer decoder is the same for both HCFormer and Mask2Former), as shown in Tab. \ref{tab:m2f-wo-ms}.
\begin{table}[t]
    \centering
    \begin{tabular}{c|cc}
           & Mask2Former~\cite{m2f} & HCFormer+\\ \shline
        PQ & 50.2 & 50.2 \\
        FLOPs & 213G & 96G
    \end{tabular}
    \caption{Comparison between Mask2Former \textit{without} using multi-scale feature maps in the transformer decoder and HCFormer+ on the COCO dataset. We use ResNet-50 as the backbone architecture.}
    \label{tab:m2f-wo-ms}
\end{table}
This result indicates HCFormer is not inferior to Mask2Former, and the gap of PQ would be due to the transformer decoder design.

In addition, we report PQ of Mask2Former replacing the well-designed pixel decoder with a simple pixel decoder, FPN~\cite{fpn}, in Tab. \ref{tab:fpn-m2f}.
\begin{table}[t]
    \centering
    \begin{tabular}{c|cc}
           & Mask2Former~\cite{m2f} & HCFormer+\\ \shline
        PQ & 50.7 & 50.2 \\
        FLOPs & 226G & 96G
    \end{tabular}
    \caption{Comparison between Mask2Former with FPN~\cite{fpn} as the pixel decoder and HCFormer+ on the COCO dataset. We use ResNet-50 as the backbone architecture.}
    \label{tab:fpn-m2f}
\end{table}
FPN is one of the most simple pixel decoders used in MaskFormer~\cite{mf} and K-Net~\cite{knet}.
This replacement reduces the PQ of Mask2Former from 51.9 to 50.7.

As seen in Tabs. \ref{tab:m2f-wo-ms} and \ref{tab:fpn-m2f}, the pixel decoder contributes the improvement of PQ.
There are some effective pixel decoder architectures and techniques leveraging the pixel decoder (\eg, the use of multi-scale feature maps as in Mask2Former) to improve the accuracy.
Thus, removing the pixel decoder makes the segmentation pipeline simple and interpretable but may lose some means to improve the segmentation accuracy, which is one of the limitations of HCFormer.

\section{Detailed experimental setup}
\label{sec:detailed-setup}
We trained models with the Swin-L backbones for the COCO dataset on 4$\times$ NVIDIA A100 GPUs and the other models on 2$\times$ NVIDIA RTX8000 GPUs.
The training protocol follows Mask2Former~\cite{m2f} except for the number of training epochs for HCFormer+.
We describe the details as follows.

\subsection{Panoptic and instance segmentation on COCO}
\label{sec:train-proto-pan}
For HCFormer with the ResNet-50 and Swin-S backbones, we set the number of training epochs and the trainable queries that are fed into the transformer decoder to 50 and 100, respectively.
For HCFormer with the Swin-L backbone, these parameters are set to 100 and 200.
For HCFormer+, the number of epochs is set to 100 for the ResNet-50 and Swin-S backbones and 200 for the Swin-L backbone, and other parameters are the same as HCFormer.
We use an initial learning rate of 0.0001 and a weight decay of 0.05 for all backbones.
A learning rate multiplier of 0.1 is applied to the backbone, and we decay the learning rate at 0.9 and 0.95 fractions of the total number of training steps by a factor of 10.
We use AdamW optimizer~\cite{adamW} with a batch size of 16.
We initialize the scale parameter $s^{(i)}$ with 0.1.
For data augmentation, we use the same policy as in Mask2Former~\cite{m2f}.
For inference, we use the Mask R-CNN inference setting~\cite{mrcnn}, where we resize an image with a shorter side to 800 and a longer side up to 1333.

\subsection{Semantic segmentation on AD20K and Cityscapes}
We use AdamW~\cite{adamW} and the poly learning rate schedule with an initial learning rate of 0.0001 and a weight decay of 0.05.
A learning rate multiplier of 0.1 is applied to the backbones.
A batch size is set to 16.
We initialize the scale parameter $s^{(i)}$ with 0.1.
For data augmentation, we use random scale jittering between 0.5 and 2.0, random horizontal flipping, random cropping, and random color jittering.
For the ADE20K dataset, if not stated otherwise, we use a crop size of 512$\times$512 and train HCFormer for 160k iterations and HCFormer+ for 320k iterations.
For the Cityscapes dataset, we use a crop size of 512$\times$1024 and train HCFormer for 90k iterations and HCFormer+ for 180k iterations.
The number of trainable queries is set to 100 for all models and both datasets.

\section{Post-processing for the MaskFormer's segmentation head and the relation to the clustering}
\label{sec:post-process}
Let $P\in\Delta^{N_m\times K+1}$ be the $(K+1)$ class probability for masks computed by eq. \eqref{eq:mask}.
For a post-processing, we assign a pixel at $j=[h,w]$ to one of the $N_m$ predicted probability-assignment pairs via $\arg\max_{i:c_i\neq\varnothing}P_{i,c_i}M_{j,i}^{(0)}$, where $\varnothing$ denotes the \texttt{no object} class and $c_i$ is the most likely class label, $c_i=\arg\max_{c\in\{1,\dots,K,\varnothing\}}P_{i,c}$, for each probability-assignment pair $i$.
Note that $M^{(0)}$ denotes the decoded masks (\ie, $M^{(0)}=\prod_iA^{(i)}M^{(5)}$).

The maximization problem in post-processing, $\arg\max_{i:c_i\neq\varnothing}P_{i,c_i}M_{j,i}^{(0)}$, is equivalent to the clustering problem, eq. \eqref{eq:clst-as-attn}.
Specifically, $P_{i,c_i}M_{j,i}$ is a similarity between the $i$-th mask query and the pixel at $j=[h,w]$, and the maximization problem selects the most similar mask query.
This procedure is the clustering problem for pixels with the mask queries as a prototype.
Thus, we can view MaskFormer's segmentation head as one of the clustering modules.

\section{Additional results}
\label{sec:add-res}
\subsection{Per-pixel classification with hierarchical clustering}
The per-pixel classification head, which is the segmentation head used in many semantic segmentation models, can also be viewed from the clustering perspective.
In this head, the weight of the linear classifier with the softmax activation that produces class probability can be viewed as a set of class prototypes.
The linear classifier groups pixels based on the class prototypes with the inner product as the similarity.
Specifically, let $w\in\mathbb{R}^{C\times K}$ be a weight matrix of the linear classifier, which corresponds to the $K$-class prototypes with $C$-dimensional features, and let $f\in\mathbb{R}^{C\times M}$ be a feature map with $M$ pixels.
Then, the per-pixel classification models classify pixels as $\mathrm{Softmax}_\text{row}(f^\top w)$, which is the same formula as the attention in eq. \eqref{eq:attn} (the norm of $w$ can be viewed as the inverse of the scale parameter $s$).
Therefore, we can also view this as the soft clustering problem and naturally incorporate our hierarchical clustering scheme into the per-pixel classification models.

We evaluate our method using the per-pixel classification model.
As a baseline, we use FCN-32s~\cite{fpn}, FPN~\cite{fpn}, PSPNet~\cite{pspnet}, and Deeplabv3~\cite{deeplabv3}, and we combine the proposed module with FCN-32s, PSPNet, and Deeplabv3.
We refer to the combined models as HC-FCN-32s, HC-PSPNet, and HC-Deeplabv3, respectively.
We use ResNet-101~\cite{resnet} as the backbone for all models.
Note that PSPNet and Deeplabv3 use dilated convolution~\cite{yu2015multi,deeplab} in the backbone, and their output stride is 8, meaning that the number of downsampling layers in the backbone is 3.
HC-PSPNet and HC-Deeplabv3 use the same backbone architecture as FCN-32s; their output stride is 32.
The training protocol follows \cite{pspnet}.
We set the crop size for ADE20K to 512$\times$512.

\begin{table}[t]
    \centering
    \begin{subtable}[a]{0.45\textwidth}
    \begin{tabular}{c|cc|c}
        Models & mIoU & Pixel Acc. & msec/image\\ \shline
        PSPNet & 42.7 & 80.9 & 48.7\\ \hdashline
        HC-PSPNet & 42.6 & 80.9 & 19.0 \\ \hline
        Deeplabv3 & 42.7 & 81.0 & 62.5\\ \hdashline
        HC-Deeplabv3 & 42.8 & 81.1 & 22.5\\
    \end{tabular}
    \caption{ADE20K}
    \end{subtable}
    \hfill
    \begin{subtable}[b]{0.45\textwidth}
    \begin{tabular}{c|cc|c}
        Method & mIoU & Pixel acc. & msec/image \\ \shline
        FCN-32s & 71.7 & 94.9 & 71.7 \\
        FPN & 75.1 & 95.8 & 82.0 \\ \hdashline
        HC-FCN-32s & 76.1 & 96.1 & 87.2 \\ \hline
        PSPNet & 77.7 & 96.2 & 298.7 \\ \hdashline
        HC-PSPNet & 77.6 & 96.2 & 88.1 \\ \hline
        Deeplabv3 & 78.4 & 96.3 & 380.8 \\ \hdashline
        HC-Deeplabv3 & 77.6 & 96.3 & 93.9
    \end{tabular}
    \caption{Cityscapes}
    \end{subtable}
    \caption{Evaluation results on semantic segmentation with per-pixel models. The inference time is measured with NVIDIA Quadro RTX8000.}
    \label{tab:ss-per-pixel}
\end{table}
We show the evaluation results for the validation set in Tab. \ref{tab:ss-per-pixel}.
The latency of HC-PSPNet and HC-Deeplabv3 is lower than that of PSPNet and Deeplabv3, and HC-PSPNet shows a comparable result for PSPNet for both datasets.
However, mIoU of HC-Deeplabv3 on Cityscapes is lower than that of Deeplabv3.
Deeplabv3 uses atrous spatial pyramid pooling (ASPP) that uses several dilated (atrous) convolution layers with different dilation, and the maximum dilation is 24, which corresponds to the convolution with a kernel size of 49.
For HC-Deeplabv3, the resolution of the feature map fed into the ASPP layer would be quite low.
As a result, HC-Deeplabv3 degrades mIoU from Deeplabv3.

Compared to FCN-32s, HC-FCN-32s significantly improves mIoU because it can preserve detailed information, such as object boundaries, via the clustering module.
In addition, HC-FCN-32s shows a superior mIoU than FPN, that is, FCN-32s with the simple pixel decoder, which is also used in MaskFormer~\cite{mf}, while the latency is higher than FPN.
This result is consistent with the results of the comparison between HCFormer and MaskFormer.

\subsection{Effect of downsampling modules}
\label{sec:app-ds}
In the experiments, we use the deformable convolution v2 (DCNv2)~\cite{dcnv2} as the downsampling for the ResNet backbone.
We verify its effect by comparing the ResNet backbone with and without DCNv2.
We train models using the same training protocol for both models with and without DCNv2 (Sec. \ref{sec:train-proto-pan}).

\begin{figure*}
    \centering
    \includegraphics[clip,width=0.8\hsize]{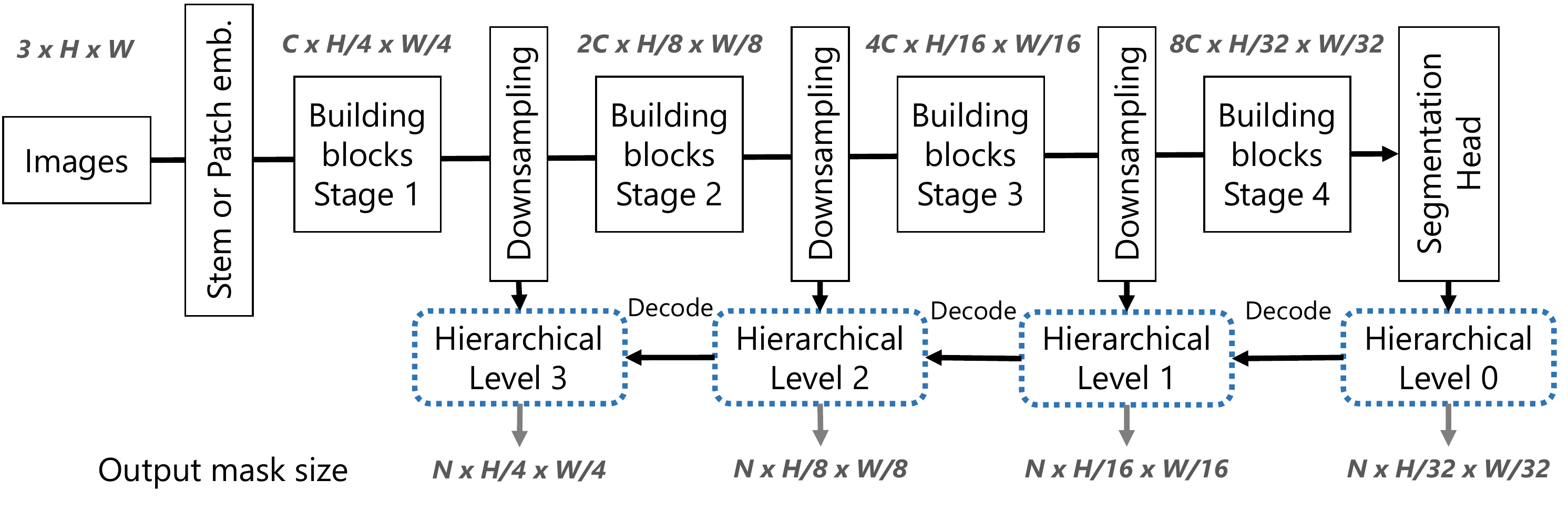}
    \caption{Backbone architecture with hierarchical clustering.}
    \label{fig:backbone}
\end{figure*}

We show the evaluation results for COCO~\cite{coco} in Tab. \ref{tab:w-w/o-dcn}.
Note that the hierarchical level is shown in Fig. \ref{fig:backbone}.
Since the downsampling layer is replaced with DCNv2 only for the layers to which the clustering module is attached, both results for the hierarchical level of 0 are the same.
DCNv2 significantly improves PQ, especially for hierarchical levels of 2 and 3.
The higher the hierarchical level, the more sampling error is accumulated.
Thus, the improvement is larger for the higher level.
\begin{table*}[h]
    \centering
    \begin{tabular}{cc}
    \scalebox{0.9}{
    \begin{tabular}{c|cccc}
        Hierarchical Level & 0 & 1 & 2 & 3 \\ \shline
        PQ & 41.9 & 45.6 & 46.7 & 47.7 \\
        FLOPs & 82G & 83G & 84G & 87G \\
        \#Params & 37M & 38M & 38M & 38M 
    \end{tabular}
    }
    &
    \scalebox{0.9}{
    \begin{tabular}{c|ccccc}
        Hierarchical Level & 0 & 1 & 2 & 3 \\ \shline
        PQ & 41.9 & 44.0 & 44.8 & 45.3 \\
        FLOPs & 82G & 83G & 84G & 87G \\
        \#Params & 37M & 38M & 38M & 38M 
    \end{tabular}
    }\\
    (a) w/ DCNv2 & (b) w/o DCNv2 
    \end{tabular}
    \caption{Evaluation results on COCO \texttt{val.} with various hierarchical levels.}
    \label{tab:w-w/o-dcn}
\end{table*}

Note that we do \textit{not} use DCNv2 as downsampling for the transformer-based backbone (Swin-S and Swin-L), and the accuracy of HCFormer with the transformer-based backbone is higher than that of MaskFormer.
Thus, the effectiveness of HCFormer is significant, although its accuracy of HCFormer with the CNN-based backbone is almost the same as that of MaskFormer.

\subsection{Additional ablation study}
\label{sec:app-ablation}
We show the additional ablation study.
Basically, we use ResNet-50 as a backbone model, except for the result in Tab. \ref{tab:swinS-hl}, and the training protocol is the same as described in Sec. \ref{sec:detailed-setup}.

We evaluate HCFormer with the transformer decoder used in MaskFormer~\cite{mf}, although the masked transformer decoder~\cite{m2f} was used in the experiments in the main manuscript.
As shown in Tab. \ref{tab:std-dec}, HCFormer with the standard transformer decoder still outperforms MaskFormer even though the DCNv2 is not used.
The difference between HCFormer w/o DCNv2 and MaskFormer in Tab. \ref{tab:std-dec} is the segmentation process: MaskFormer generates segmentation masks directly from the per-pixel features, and HCFormer generates them based on the hierarchical strategy.
This indicates that our hierarchical clustering scheme generates more accurate masks than direct inference from the per-pixel features.
\begin{table}[t]
    \centering
    \begin{tabular}{c|cc}
        method & PQ & FLOPs\\ \shline
        MaskFormer~\cite{mf} & 46.5 & 181G \\
        Mask2Former~\cite{m2f} & 47.1 & 213G \\ \hdashline
        HCFormer w/o DCNv2 & 46.8 & 97G\\
        HCFormer w/ DCNv2 & 47.6 & 97G \\
    \end{tabular}
    \caption{Evaluation results of HCFormer with the standard transformer decoder used in ~\cite{detr,mf}. Note that Mask2Former does not use masked attention but uses multi-scale feature maps in the transformer decoder in this comparison.}
    \label{tab:std-dec}
\end{table}
We also show the results of Mask2Former without the masked attention in the transformer decoder.
PQ of HCFormer without DCNv2 is slightly worse than that of Mask2Former, but we believe that due to the use of multi-scale feature maps, as described in Sec. \ref{sec:diff-m2f}.

\begin{table}[t]
    \centering
    \begin{tabular}{c|ccccc}
        \#Queries         & 20 & 50 & 100 & 150 & 200 \\ \shline
        COCO (PQ)         & 42.2 & 46.3 & 47.7 & 47.9 & 47.9\\ 
        ADE20K (mIoU)     & 44.4 & 45.5 & 45.5 & 44.8 & 44.4\\
        Cityscapes (mIoU) & 75.9 & 75.4 & 76.7 & 76.2 & 76.4
    \end{tabular}
    \caption{Evaluation results with various numbers of trainable queries.}
    \label{tab:n-queries}
\end{table}
\begin{table}[t]
    \begin{tabular}{c|cccc}
        Hierarchical Level & 0 & 1 & 2 & 3 \\ \shline
        mIoU     & 42.2 & 43.4 & 44.0 & 45.5 \\
        FLOPs    & 28G & 28G & 28G & 29G \\
        \#Params & 37M & 38M & 38M & 38M
    \end{tabular}
    \caption{Evaluation results on ADE20K \texttt{val.} with various hierarchical levels.}
    \label{tab:ss-hl}
\end{table}
\begin{table}[t]
    \begin{tabular}{c|cccc}
        Hierarchical Level & 0 & 1 & 2 & 3 \\ \shline
        PQ       & 47.5 & 49.7 & 50.5& 50.9 \\
        FLOPs    & 167G & 167G & 168G & 170G \\
        \#Params & 62M & 62M & 62M & 62M 
    \end{tabular}
    \caption{Evaluation results on COCO \texttt{val.} with various hierarchical levels for the Swin-S backbone.}
    \label{tab:swinS-hl}
\end{table}
\begin{table}[t]
    \begin{tabular}{c|ccccc}
        \#Decoder Layers & 1 & 2 & 4 & 8 & 16\\ \shline
        mIoU & 43.2 & 44.4 & 44.5 & 45.5 & 46.1\\
        FLOPs & 28G & 28G & 28G & 29G & 31G\\
        \#Params & 27M & 29M & 32M & 38M & 51M
    \end{tabular}
    \caption{Evaluation results on ADE20K \texttt{val.} with various numbers of transformer decoder layers.}
    \label{tab:num-dlayer-ade}
\end{table}
\begin{figure*}[t]
    \centering
    \begin{tabular}{ccc|c}
        \includegraphics[clip,width=0.23\hsize]{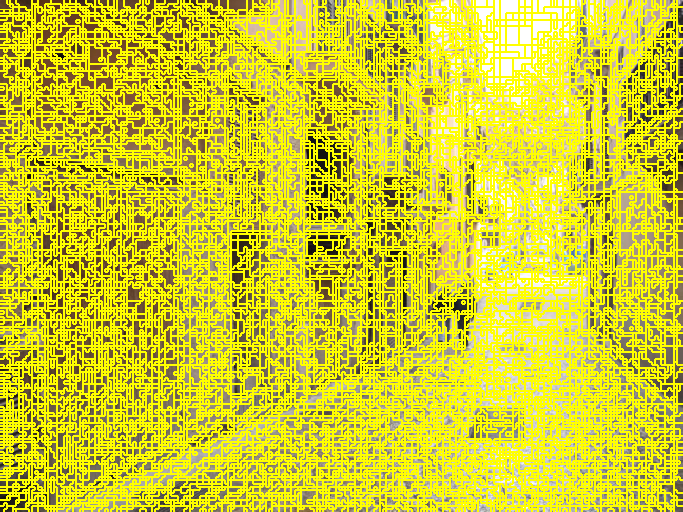} &
        \includegraphics[clip,width=0.23\hsize]{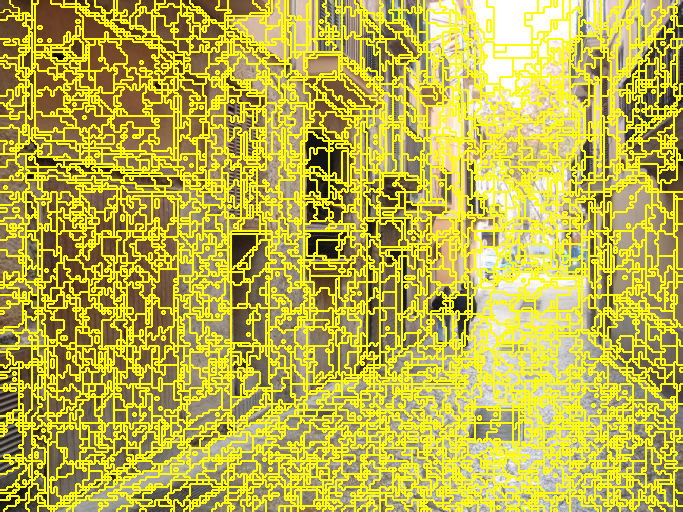} &
        \includegraphics[clip,width=0.23\hsize]{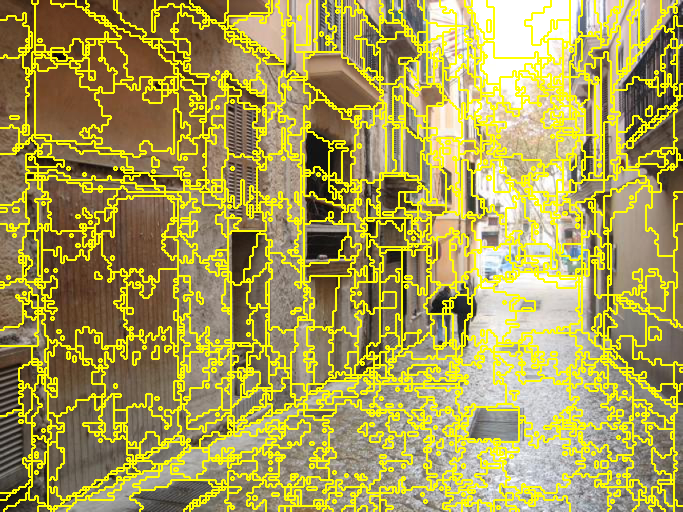} &
        \includegraphics[clip,width=0.23\hsize]{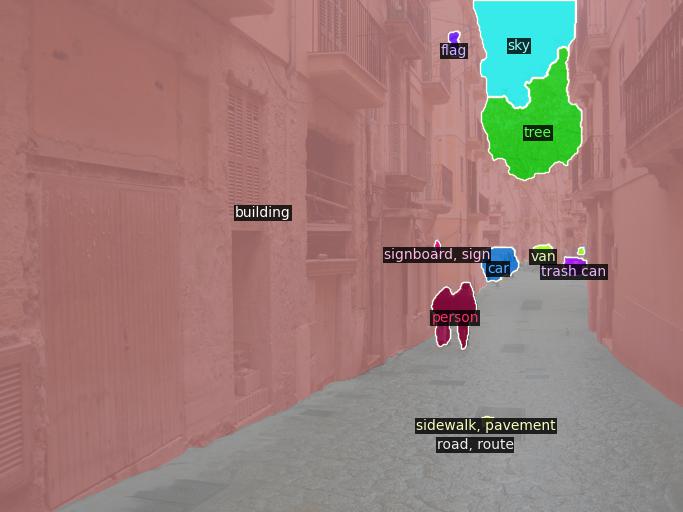}\\
        \includegraphics[clip,width=0.23\hsize]{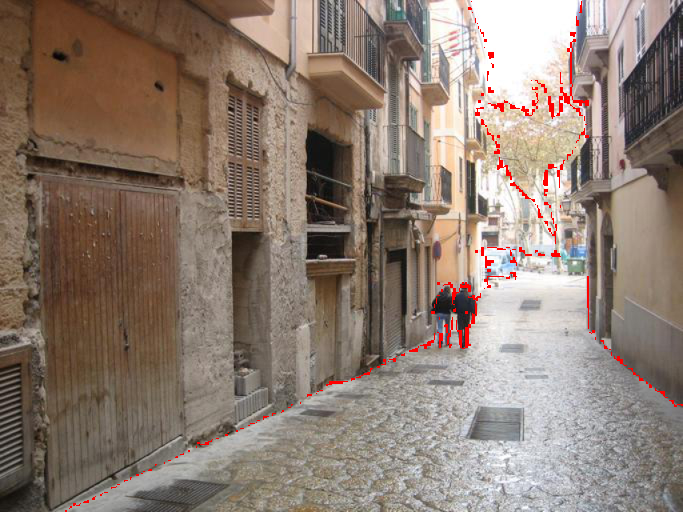} &
        \includegraphics[clip,width=0.23\hsize]{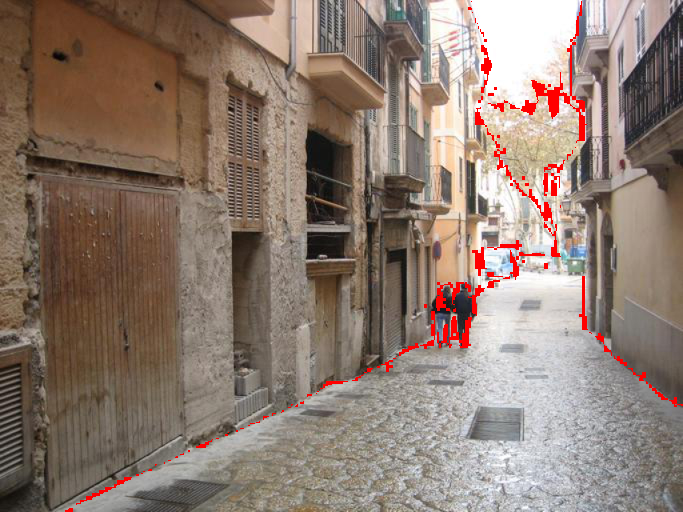} &
        \includegraphics[clip,width=0.23\hsize]{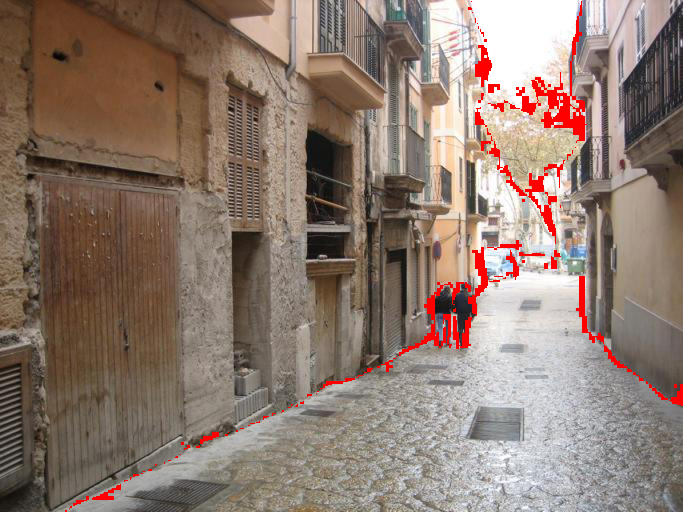} &
        \includegraphics[clip,width=0.23\hsize]{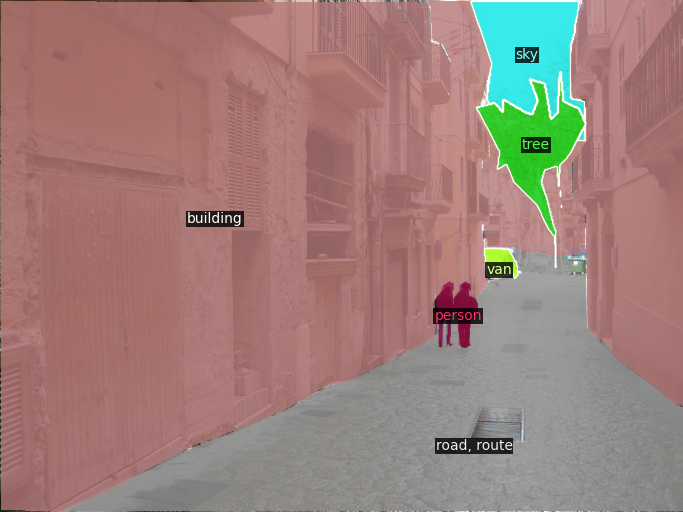} \\ \hline
        \includegraphics[clip,width=0.23\hsize]{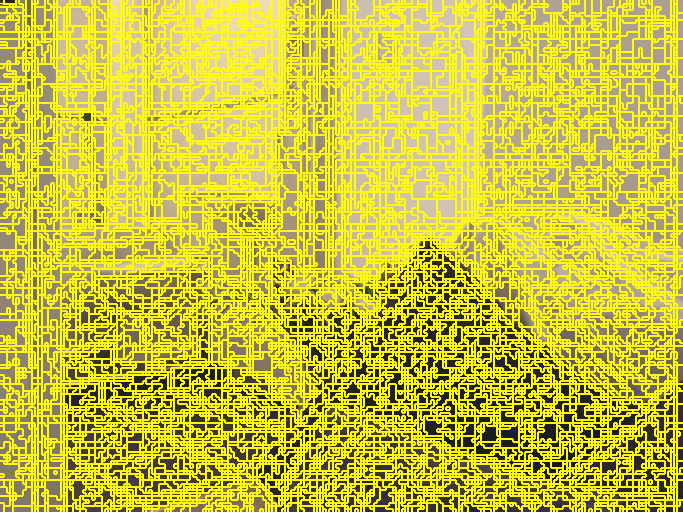} &
        \includegraphics[clip,width=0.23\hsize]{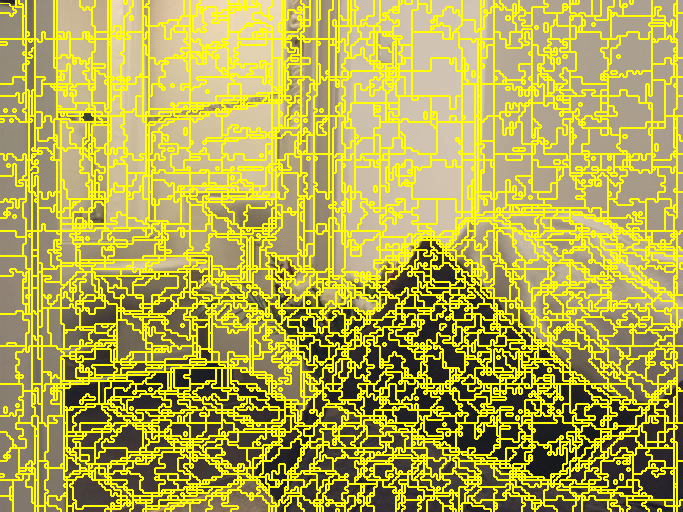} &
        \includegraphics[clip,width=0.23\hsize]{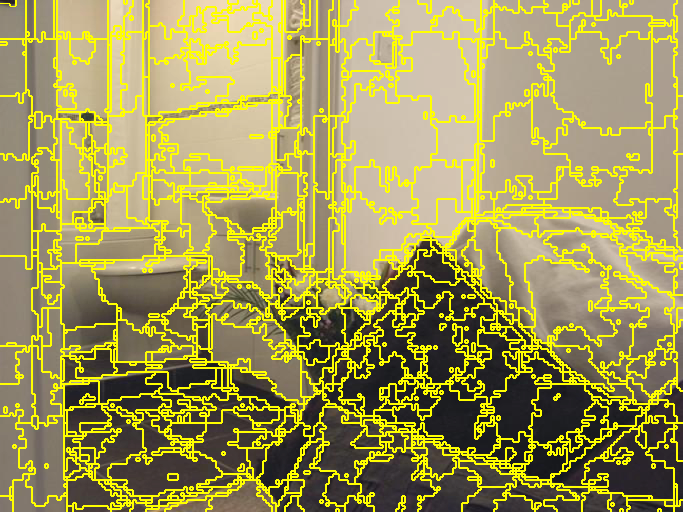} &
        \includegraphics[clip,width=0.23\hsize]{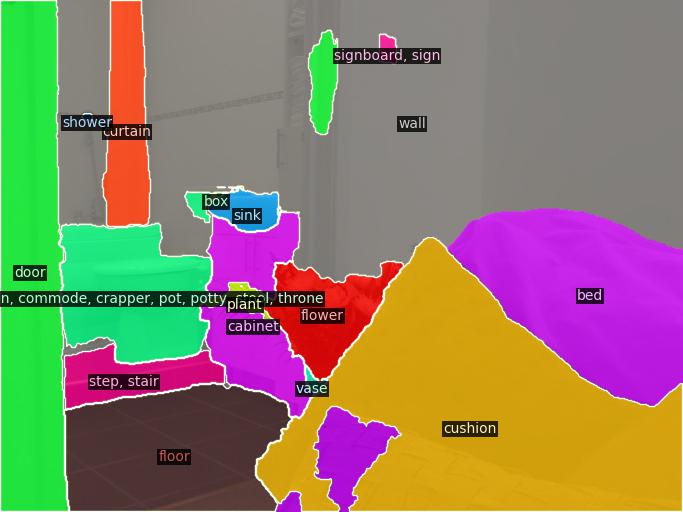}\\
        \includegraphics[clip,width=0.23\hsize]{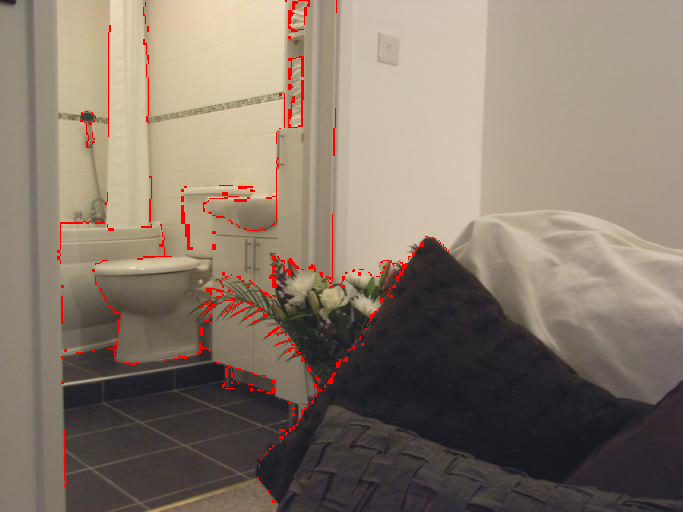} &
        \includegraphics[clip,width=0.23\hsize]{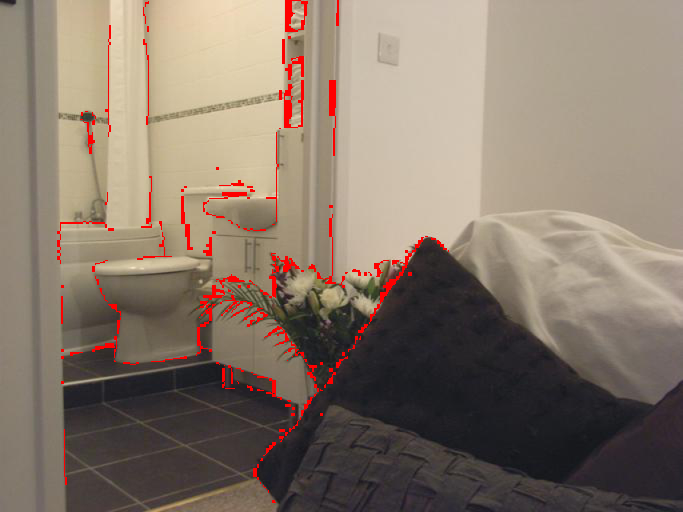} &
        \includegraphics[clip,width=0.23\hsize]{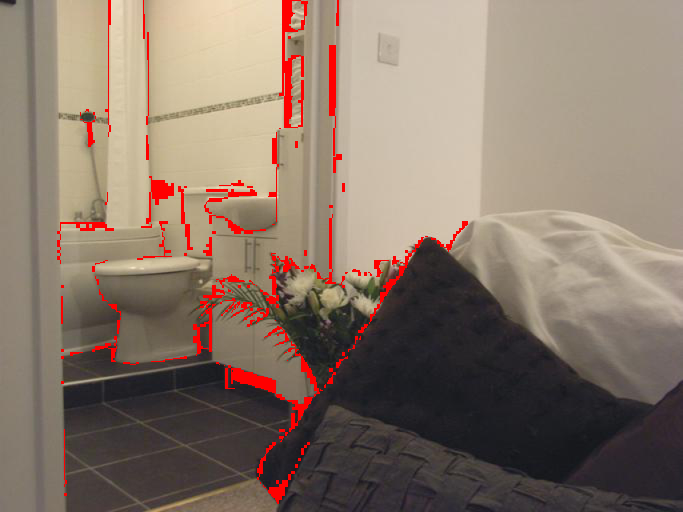} &
        \includegraphics[clip,width=0.23\hsize]{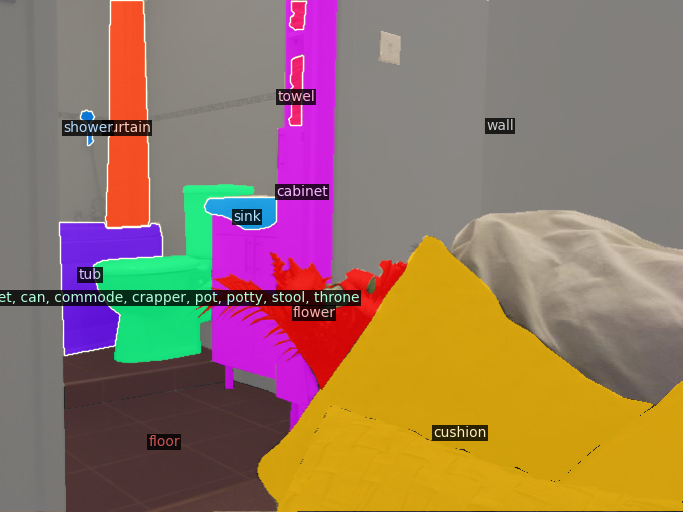} \\ 
    \end{tabular}
    \caption{Example results on ADE20K. The top row shows the cluster boundaries and predicted masks, and the bottom row shows the undersegmentation error~\cite{sp-sota} as red regions and the ground-truth label.}
    \label{fig:ade-vis}
\end{figure*}
\begin{figure*}[t]
    \centering
    \begin{tabular}{ccc|c}
        \includegraphics[clip,width=0.23\hsize]{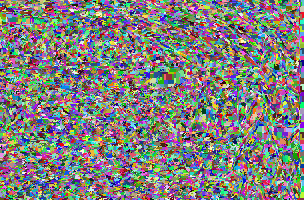} &
        \includegraphics[clip,width=0.23\hsize]{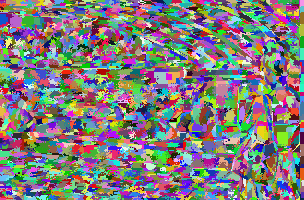} &
        \includegraphics[clip,width=0.23\hsize]{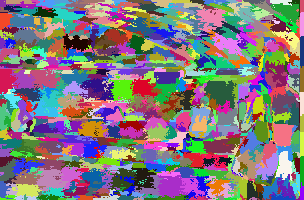} &
        \includegraphics[clip,width=0.23\hsize]{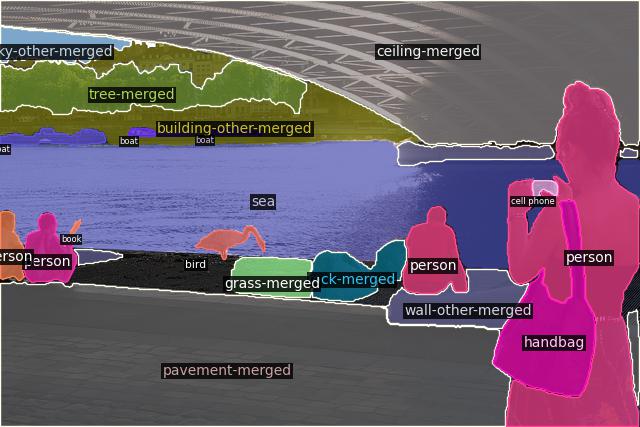} \\
        \includegraphics[clip,width=0.23\hsize]{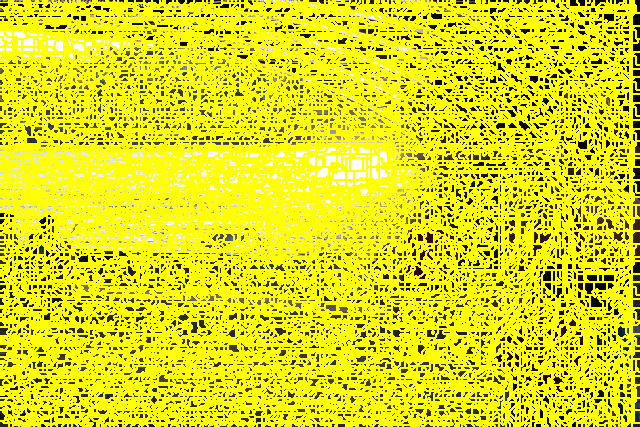} &
        \includegraphics[clip,width=0.23\hsize]{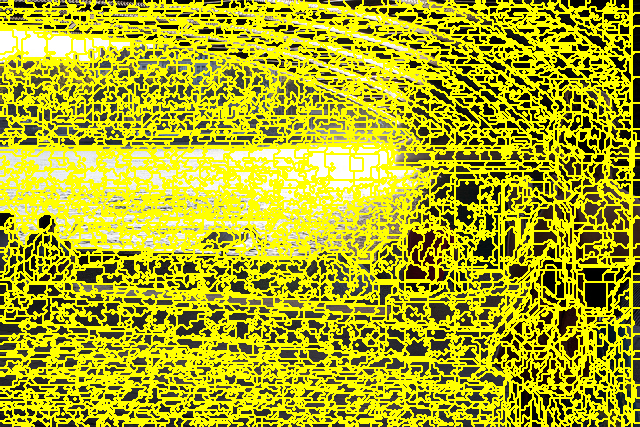} &
        \includegraphics[clip,width=0.23\hsize]{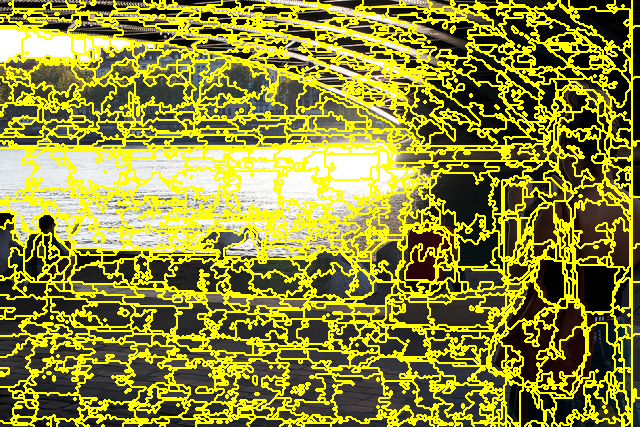} &
        \includegraphics[clip,width=0.23\hsize]{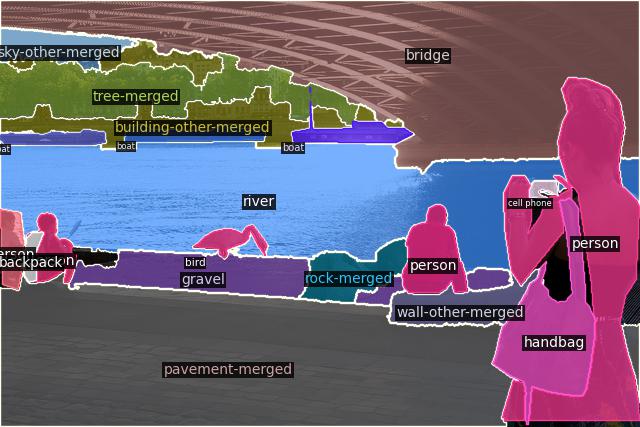} \\ \hline
        \includegraphics[clip,width=0.23\hsize]{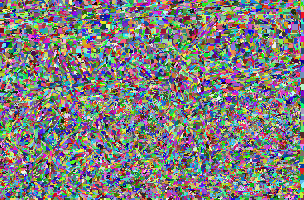} &
        \includegraphics[clip,width=0.23\hsize]{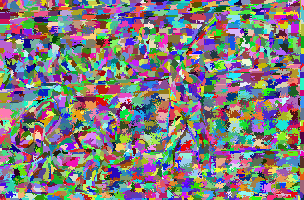} &
        \includegraphics[clip,width=0.23\hsize]{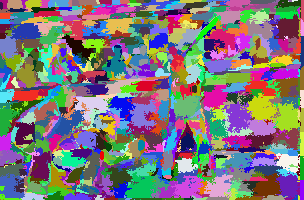} &
        \includegraphics[clip,width=0.23\hsize]{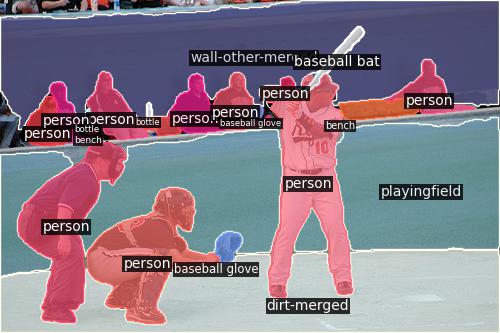} \\
        \includegraphics[clip,width=0.23\hsize]{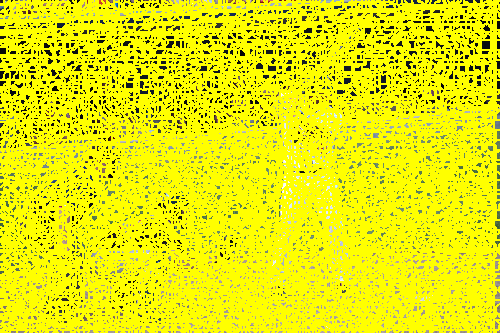} &
        \includegraphics[clip,width=0.23\hsize]{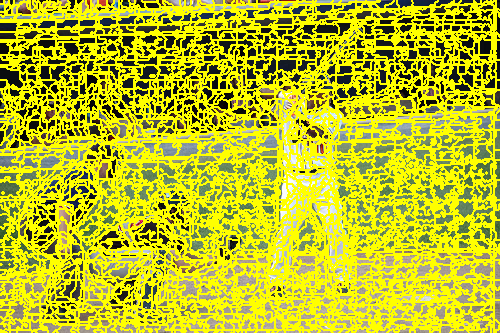} &
        \includegraphics[clip,width=0.23\hsize]{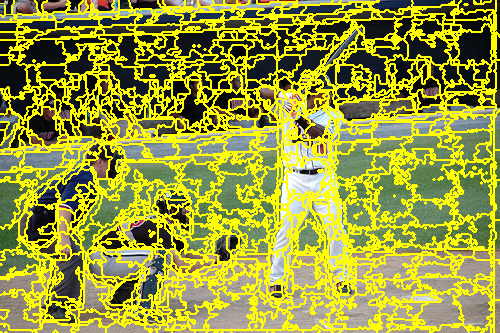} &
        \includegraphics[clip,width=0.23\hsize]{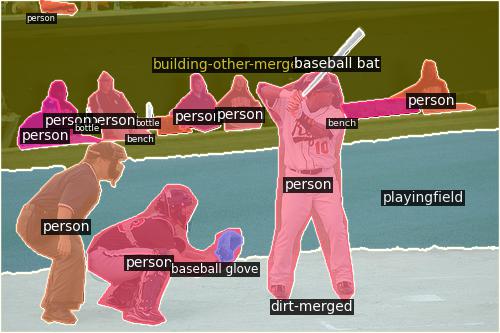} \\
    \end{tabular}
    \caption{Example results on COCO. The top row shows hierarchical clustering results to which a random color is assigned and prediction, and the bottom row shows cluster boundaries and ground truth.}
    \label{fig:coco-vis}
\end{figure*}

We evaluate HCFormer using a various number of trainable queries (Tab. \ref{tab:n-queries}).
The segmentation accuracy is saturated at about 100 or 150.
For memory efficiency, 50 or 100 queries would be preferred.

We evaluate HCFormer with the various hierarchical levels on the semantic segmentation (ADE20K).
We verify the improvement of the accuracy on the semantic segmentation, as shown in Tab. \ref{tab:ss-hl}.
In addition, we evaluate HCFormer with the Swin-S~\cite{swin} backbone with various numbers of hierarchical levels on COCO~\cite{coco}.
Regardless of the backbone type and dataset, a higher hierarchical level leads to higher accuracy.

We evaluate HCFormer with the ResNet-50 backbone with various numbers of transformer decoder layers on ADE20K.
MaskFormer~\cite{mf} reports reasonable semantic segmentation performance (43.0 mIoU on ADE20K) with only one transformer decoder layer.
HCFormer with only one transformer decoder layer also shows reasonable results (43.2 mIoU on ADE20K), as shown in Tab. \ref{tab:num-dlayer-ade}.
This result is better than the per-pixel classification baselines with ResNet-101~\cite{resnet}, such as PSPNet~\cite{pspnet} and Deeplabv3~\cite{deeplabv3}, as shown in Tab. \ref{tab:ss-per-pixel}.
The deeper transformer decoder improves mIoU, and four or eight layers would be sufficient for HCFormer.

\subsection{Visualization}
\label{sec:app-vis}
We show example results on ADE20K by HCFormer with Swin-L~\cite{swin} in Fig. \ref{fig:ade-vis}.
Basically, the undersegmentation error appears near the boundaries because of ambiguity and downsampling in stem and patch embedding layers of the backbone to which the clustering module is not attached.
In the outdoor image (top row), the undersegmentation error appears in the tree region (green region) at the coarser clustering levels, but this would be due to ambiguity in the annotation.
In the indoor image (bottom row), the undersegmentation error appears below the cabinet region (pink region) in all the levels because the pixels in the wall and floor regions are grouped.
In addition, HCFormer misclassifies the cabinet and the towel as wall and door, respectively.
The backbone model would not recognize their semantics, although it was able to discriminate between pixels in the cabinet and pixels in the wall because the undersegmentation error only appears in their boundaries.

We show example results for COCO panoptic segmentation by HCFormer+ with Swin-L~\cite{swin} in Fig. \ref{fig:coco-vis}.
HCFormer groups pixels in images well, although there are some misclassifications.
To solve such a misclassification error, we might need to improve the transformer decoder and/or incorporate stronger backbones.

\end{document}